\documentclass[review,11pt]{JMtemplate}
\usepackage{bm}
\usepackage{graphicx} 
\usepackage{amsmath,mathrsfs,amssymb} 
\usepackage{amsfonts}  
\usepackage{algorithm}
\usepackage{algorithmic}
\usepackage{siunitx}   
\usepackage{upgreek}   
\usepackage{xcolor}    
\usepackage{wrapfig}
\usepackage{subfigure}
\usepackage{booktabs}       
\usepackage{longtable}
\usepackage{threeparttable}
\usepackage{multirow}
\usepackage{multicol}
\usepackage{times}
\usepackage{helvet}
\usepackage{courier}
\usepackage{makecell}   
\usepackage[hyphens]{url} 
\usepackage{natbib}  
\usepackage{caption} 

\newtheorem{theorem}{Theorem}

\newtheorem{definition}{Definition}

\newtheorem{assumption}{Assumption}

\newcommand*{\dif}{\mathop{}\!\mathrm{d}}
\newcommand*{\tr}{\mathop{}\!\mathrm{tr}}
\newcommand*{\diag}{\mathop{}\!\mathrm{diag}}
\newcommand*{\re}{\mathop{}\!\mathrm{Re}}
\newcommand*{\im}{\mathop{}\!\mathrm{Im}}
\newcommand*{\e}{\mathop{}\!\mathrm{e}}
\newcommand*{\ii}{\mathop{}\!\mathrm{i}}
\newcommand*{\op}{\mathop{}\!\mathrm{o}_{\mathrm{P}}}

\begin{document}
	
\begin{frontmatter}
	
\title{ARISE: ApeRIodic SEmi-parametric Process for Efficient Markets without Periodogram and Gaussianity Assumptions}

\author{Shao-Qun Zhang}
\author{Zhi-Hua Zhou\footnote{Zhi-Hua Zhou is the corresponding author.}}
\address{National Key Laboratory for Novel Software Technology\\
	Nanjing University, Nanjing 210093, China\\
    \{zhangsq,zhouzh\}@lamda.nju.edu.cn }
\date{\today}


\begin{abstract}
Mimicking and learning the long-term memory of efficient markets is a fundamental problem in the interaction between machine learning and financial economics to sequential data. Despite the prominence of this issue, current treatments either remain largely limited to heuristic techniques or rely significantly on periodogram or Gaussianty assumptions. In this paper, we present the \textit{ApeRIodic SEmi-parametric} (ARISE) process for investigating efficient markets. The ARISE process is formulated as an infinite-sum function of some known processes and employs the aperiodic spectrum estimation to determine the key hyper-parameters, thus possessing the power and potential of modeling the price data with long-term memory, non-stationarity, and aperiodic spectrum. We further theoretically show that the ARISE process has the mean-square convergence, consistency, and asymptotic normality without periodogram and Gaussianity assumptions. In practice, we apply the ARISE process to identify the efficiency of real-world markets. Besides, we also provide two alternative ARISE applications: studying the long-term memorability of various machine-learning models and developing a latent state-space model for inference and forecasting of time series. The numerical experiments confirm the superiority of our proposed approaches.	
\end{abstract}


\end{frontmatter}

\section{Introduction} \label{sec:introduction}
Machine learning has become a hotwave in financial economics due to its advances in handling complex sequential data~\citep{hyndman2020,sezer2020:stock}. An omnipresent challenge in these sequence modeling tasks is to represent and learn the serial patterns of asset markets. From the efficient market theory~\citep{fama1970:EMH,schwert2003}, stock prices usually exhibit the characteristics of long-term memory, non-stationarity, and aperiodicity. In machine learning, there are a lot of theoretical and practical works~\citep{kuznetsov2015:pac,zhang2017:stock} to handle non-stationary and aperiodic data, whereas lacking consolidated specification on whether a given machine learning model has captured the long-term memory patterns of efficient markets~\citep{bengio1994,greaves2019:SG}. Thus, developing machine learning with the complementary ability to model efficient markets has become a desirable but challenging issue.


The past decades have witnessed tremendous efforts on this issue. The most famous are statistical approaches, such as the \textit{Auto-Regressive-Moving-Average} (ARMA) and generalized auto-regressive conditional heteroskedasticity models~\citep{box2015}. Analysts usually employ the auto-correlation to measure the long-term memory or dependency between the prices at different timestamps. However, such statistical models cannot work only beyond stationarity assumptions and the linear auto-correlation coefficient is truncate-tailed, i.e., decays exponentially as time evolves, limiting the applicability of such methods. Alternative approaches are based on deep learning, such as RNN, LSTM~\citep{hochreiter1997}, and their variants~\citep{cho2014,zhou2016}, which deliver laudable handling non-stationary and aperiodic time series. Nevertheless, the deep learning models often lack comprehensibility and make the whole model black boxes. Furthermore, it is still a highly controversial issue whether and to what extent a concerned deep learning model has learned to represent long-term memory between observations.

To model the stock data with long-term memory, some investigators focus on the semi-parametric treatment of the fractionally integrated process, which has a temporal representation that naturally enhances conventional ARMA models with non-integer values of the differencing parameter~\citep{beran1994,kim2008}. A positive differencing parameter explicitly induces an infinite sum of the long-range time series, thus allowing for mimicking and identifying the long-term memory of efficient markets in a statistical sense. Unfortunately, the existing approaches are heavily reliant on the periodogram or Gaussianity assumptions and periodogram estimation~\citep{greaves2019:SG,shimotsu2007:GSE}, which necessitates strong distributional and regularity conditions and prevents the extraction of the aperiodic spectrum and non-stationary patterns.

In this paper, we propose the \textit{ApeRIodic SEmi-parametric Process} (ARISE) process for investigating efficient markets from the statistical perspective. The ARISE process is a semi-parametric approach that combines a parametric integrated model with a non-parametric \textit{Aperiodic Spectrum Estimation} (ASE). The parametric integrated model is formulated as an infinite-sum function of some known processes, with the potential of modeling the efficient markets with long-term memory. The non-parametric wavelet-threshold approach is used for aperiodic spectrum estimation, giving it the ability to handle non-stationary and aperiodic signals. To show the well-posedness of our proposed ARISE process, we also theoretically establish the fundamental properties, i.e., the mean-square convergence with near-optimal rate, consistency, and asymptotic normality, of the ASE without assuming periodogram and Gaussianity. We apply the ARISE process to investigate several real-world markets and confirm their efficiency. Furthermore, we also explore other ARISE applications to two machine-learning scenarios, that is, i) investigating the long-term memorability of several machine-learning models, ii) refining the latent state-space model for inference and forecasting of time series. The numerical experiments conducted on the simulated and real-world non-stationary data sets confirm the superiority of our proposed approaches.		

The rest of this paper is organized as follows. Section~\ref{sec:rw} reviews the related work. Section~\ref{sec:ARISE} introduces our ARISE process with a concrete implementation. Section~\ref{sec:TA} establishes the well-posed analysis of the ARISE process. Section~\ref{sec:applications} provides three applications for the ARISE process. Finally, Section~\ref{sec:conclusions} concludes our work with discussions and prospects.

\section{Related Work} \label{sec:rw}
\noindent\textbf{About the efficient market hypothesis.} The efficient market hypothesis, alternatively known as the efficient market theory~\citep{fama1970:EMH}, is one of the cornerstones in financial economics. In contrast to the random walk theory~\citep{fama1965:RWT,kendall1953:RWT} that share prices resemble a random walk and thus investors cannot predict their futures based on historical prices, the efficient market hypothesis claims two significant characteristics. First, asset markets follow a fair game, with prices established by a large number of rational investors voting at time-varying horizons~\citep{peters1994:EMH}. Second, share prices comprise all information where the efficient market's past, current, and even future events will be posted in its prices~\citep{schwert2003}. Thus, the efficient market data usually exhibits three conspicuous characteristics: long-term memory, non-stationary, and aperiodic spectrum, making it challenging to model its responses to environmental change. ~\\

\noindent\textbf{About the long-term memory in machine learning.} Representation and learning of long-term memory is a fundamental problem confronted in machine learning to sequential data. Around 1990, a group of researchers~\citep{elman1990,jordan1986} presented a connectionist framework that reuses recurrent links to model complex sequential data with dynamical memory. This framework opens the door that explores serial patterns of sequence or time-series data using recurrent connectionist models, and then more complex architectures, such as RNN~\citep{pearlmutter1995}, LSTM~\citep{hochreiter1997}, gated recurrent units~\citep{cho2014,zhou2016}, and their variants, were proposed. However, an omnipresent challenge is how to ensure a machine learning model has captured long-range dependencies between observations. The popular opinion is proposed by~\citet{bengio1994}, which consider the sensitivity of outputs on inputs via gradient chains and show that a recurrent system trained based on gradient descents is hard to achieve long-term memorability due to the notorious problems of gradient vanishing and explosion. Subsequent works often strive to overcome the gradient problem and verify their conclusions through performance comparison~\citep{arjovsky2016}, kernel methods~\citep{lei2017}, and ablation experiments~\citep{levy2018}. Alternative approaches are based on statistics and dynamic systems. For example, \citet{greaves2019:SG} provide a semi-parametric estimation for investigating the memorability of RNNs and LSTMs, and then conclude that RNNs have short memory unless inputting stationary signals. \citet{zhao2020} show that RNNs and LSTMs without exogenous input work like Markovian update dynamics~\citep{cheng2016}, thus having short memory in statistics. ~\\

\noindent\textbf{About the fractionally integrated process.} Fractionally integrated processes have a temporal representation that naturally enhances conventional ARMA models with non-integer values of the differencing parameter and are widely used as the prescriptive priors in abundant real-world fields, such as hydrology~\citep{gharari2018:hydrology,hurst1951:hydrology}, language~\citep{greaves2019:SG}, econometrics~\citep{mandelbrot2007:stock,mccauley2008:stock}, etc. The fractional differencing parameter (or equally, memory parameter) explicitly indicates the long-term dependent (i.e., persistent) or fluctuating (i.e., anti-persistent) structure of the concerned processes. The past decades have witnessed an increasing interest on the statistical estimation of the memory parameters. Fox and Taqqu~\citep{fox1986:parametric} investigated a parametric approach based on the maximum likelihood estimation at a relatively early stage. Later, Luceno~\citep{luceno1996:fast} and Tsay~\citep{tsay2010:fast} presented faster alternatives for the maximum likelihood estimation via the approximation of the quadratic form of the Gaussian likelihood function and the multivariate Durbin-Levinson algorithm, respectively. Although the parametric estimator produces promising results, it remains an expensive computational cost. Alternatively, Robinson~\citep{robinson1995gaussian} first proposed a semi-parametric estimator that consists of a likelihood-based optimization and an empirical spectral density estimator. Lobato~\citep{lobato1999:two} and Shimotsu~\citep{shimotsu2007:GSE} developed a two-step semi-parametric estimator and the \textit{Gaussian Semi-parametric Estimator} (GSE), respectively, with the rigorous treatment of consistency and asymptotic normality under periodogram or Gaussianity assumptions. Some evidences~\citep{pumi2013:evidence} show that the semi-parametric estimator can provide a more robust performance under milder conditions than the parametric one. 	

Despite promising progress, it is still immature for the statistical estimators of the memory parameter to support the theoretical analysis or applications in machine learning. First, rigorous asymptotic theory for semi-parametric estimators, including $T^{1/2}$-consistency and asymptotic normality, is heavily reliant on the periodogram or Gaussianity assumptions, which necessitates strong distributional and regularity conditions and is non-robust concerning the parametric specification of the model, leading to inconsistent estimates once mis-specified. Second, almost semi-parametric estimators employ the periodogram as the empirical spectral density estimator to construct the log-likelihood function near the zero frequency. It's well known that the periodogram presents wild fluctuations near the origin and is not a consistent estimator of the spectral density despite asymptotically unbiased. Some researchers~\citep{nielsen2011:evidence,pumi2013:evidence} use the data taper as an applied prior or the linear smoothing technique, e.g., kernel smoothing, to compute the periodogram. Unfortunately, these parametric methods cannot properly generalize the estimators of the memory parameters to non-stationary data since they are either inconsistent, requiring an apposite smoothed prior to assure consistency, or incapable of achieving the optimal mean-square rate of convergence in cases where the underlying regression function possesses a low degree of regularity. This fact not only limits the theoretical analysis of the semi-parametric estimators for the memory parameters, relying heavily on priors assumptions, but also makes the semi-parametric estimators extremely difficult to handle non-stationary signals, which significantly hinders the promotion and use of this series of work in machine learning and financial economics.

\section{Our Method} \label{sec:ARISE}
In this section, we will introduce the ARISE process. Before that, it is necessary to introduce some notations. Let $z = z_1 + z_2 \ii$ be a complex number for $z_1, z_2 \in \mathbb{R}$ and $\ii=\sqrt{-1}$ denotes the imaginary unit. We denote by $\bar{z} = z_1 - z_2 \ii$ and $|z|^2 = z_1^2 + z_2^2$. For complex-valued matrix $\mathbf{A}$, $\bar{\mathbf{A}}$ denotes the conjugate of $\mathbf{A}$. Let $\re[\cdot]$ and $\im[\cdot]$ denote the operators of extracting real and imaginary parts from a complex-valued formation, respectively, for example, $\re(z_1 + z_2 \ii) = z_1$ and $\im(z_1 + z_2 \ii) = z_2$ for $z_1, z_2 \in \mathbb{R}$. Let $[N] = \{1,2,\dots,N\}$ be the set for an integer $N > 0$ and $|\cdot|_{\#}$ denotes the number of elements in a collection, e.g., $|[N]|_{\#} = N$. Two $n$-by-$n$ matrices $\mathbf{A}$ and $\mathbf{B}$ are called \textit{similar} if there exists an invertible $n$-by-$n$ matrix $\mathbf{P}$ such that $\mathbf{B} = \mathbf{P}^{-1} \mathbf{A} \mathbf{P}$, denoted as $\mathbf{A} \sim \mathbf{B}$. Given a function $g(n)$, we denote by $h_1(n)=\Theta(g(n))$ if there exist positive constants $c_1, c_2$ and $n_0$ such that $c_1g(n) \leq h_1(n) \leq c_2g(n)$ for every $n \geq n_0$; $h_2(n)=\mathcal{O}(g(n))$ if there exist positive constants $c$ and $n_0$ such that $h_2(n) \leq cg(n)$ for every $n \geq n_0$; $h_3(n)=\Omega(g(n))$ if there exist positive constants $c$ and $n_0$ such that $h_3(n) \geq cg(n)$ for every $n \geq n_0$; $h_4(n)=o(g(n))$ if there exist positive constants $c$ and $n_0$ such that $h_4(n) < cg(n)$ for every $n \geq n_0$.

\subsection{Parametric Integrated Process} \label{subsec:MFIP}
This work considers an $l$-dimensional parametric integrated process $\{\mathbf{X}_t\}_{t=0}^{\infty}$ with $\mathbb{E}(\mathbf{X}_{it}) = 0$ for $i\in[q]$ and $\boldsymbol{d} = (d_{1}, \dots, d_{l})^{\top} \in (-1/2,1/2)^l$, generated by
\begin{equation} \label{eq:starting}
	\begin{pmatrix}
		(1-\mathfrak{B})^{d_{1}} & & 0 \\
		& \ddots & \\
		0 & & (1-\mathfrak{B})^{d_{l}}
	\end{pmatrix} \begin{pmatrix}
		\mathbf{X}_{1 t} \\
		\vdots \\
		\mathbf{X}_{l t}
	\end{pmatrix} = \begin{pmatrix}
		\epsilon_{1 t} \\
		\vdots \\
		\epsilon_{l t}
	\end{pmatrix}
\end{equation}
where $\mathfrak{B}$ is the backward-shift operator, satisfying that
$\mathfrak{B}^k\mathbf{X}_{it} = \mathbf{X}_{i(t-k)}$ for $ k\in\mathbb{N}^+$ and $i \in [l] $, and the source process $\{\boldsymbol{\epsilon}_t = (\epsilon_{1 t}, \dots, \epsilon_{l t})^{\top} \}_{t=0}^{\infty}$ is weakly stationary whose spectral density $f_{\epsilon}(\lambda)$ is bounded and bounded away from zero when frequency $\lambda$ tends to zero~\citep{box2015}. Let $\mathbf{X}_t = (\mathbf{X}_{1t}, \dots, \mathbf{X}_{lt})^{\top}$. Generally, the spectral density $f_X(\lambda)$ of $\mathbf{X}_t$ should meet two necessary requirements, i.e.,  $f_X(\lambda) >0 $ and $f_X(\lambda)$ is of finite total variation over $[-\pi,\pi]$. The former is natural for variance stabilization and the latter is a mild smoothness assumption on $f_X$. 

Provided $d_i>0$ for $i\in[l]$, one has
\begin{equation} \label{eq:B_extension}
	\begin{aligned}
		\mathbf{X}_t &= \diag_{i \in\{1, \cdots, l\}} \left\{ (1 - \mathfrak{B})^{-d_i} \right\} \boldsymbol{\epsilon}_t = \diag_{i \in\{1, \cdots, l\}} \left\{ \sum_{j=0}^{\infty} \frac{\Gamma(d_i+j)}{\Gamma(d_i) j!} \mathfrak{B}^j \right\} \boldsymbol{\epsilon}_t \\
		&= \sum_{j=0}^{\infty} \diag_{i \in\{1, \cdots, l\}} \left\{ \frac{\Gamma(d_i+j)}{\Gamma(d_i) j!} \right\} \boldsymbol{\epsilon}_{t-j} ,	
	\end{aligned}
\end{equation}
where $\Gamma(\cdot)$ is the Gamma function. It's observed that the generated process $\{ \mathbf{X}_t\}_{t=0}^{\infty}$ becomes the infinite sum of the source process $\{ \boldsymbol{\epsilon}_t\}_{t=0}^{\infty}$. Further, for $i\in[l]$, $\{\mathbf{X}_{it}\}_{t=0}^{\infty}$ with positive $d_i >0$ exhibits long-range dependence, i.e., persistent process, the auto-correlation $\gamma_k \propto k^{2d_i-1}$ as $k \to \infty$ and $f_X(\lambda) \propto \lambda^{-2d_i}$ as $\lambda \to 0^+$. In contrast, $d_i < 0$ leads to an anti-persistent process. The parametric vector $\boldsymbol{d}$ is called the \textit{memory parameter}. Therefore, the parametric integrated process, described by Eq.~\eqref{eq:starting}, \textbf{has the power and potential of mimicking the price data with long-term memory}.

The rest of this section revolves around how to calculate the memory parameter. The fundamental consensus follows the seminal works presented by~\citet{sowell1989:mle} and~\citet{robinson1995gaussian}, in which there exists a symmetric and positive-definite matrix $G \in \mathbb{R}^{l \times l}$ such that
\[
f_X(\lambda) = \Lambda(\lambda) ~f_{\epsilon}(\lambda)~ \overline{\Lambda(\lambda)} \quad\text{and}\quad f_{\epsilon}(\lambda) \sim G , \quad\text{with}\quad \Lambda(\lambda) = \diag_{i \in\{1, \cdots, l\}}\{ (1-\mathfrak{B})^{-d_i} \} .
\]
It's observed that the consensus above relies on the calculations of $\Lambda(\lambda_j)$ and $f_X(\lambda_j)$ for $j\in[m]$. To solve these issues, Shimotsu~\citep{shimotsu2007:GSE} provides a precise expansion of the operator $\overline{\Lambda(\lambda)}$ according to
\begin{equation} \label{eq:expansion}
	\left(1-e^{\ii \lambda}\right)^{d_i} = \lambda^{d_i} \e^{\ii(\lambda-\pi) d_i / 2} \left(1 + \mathcal{O}\left(\lambda^{2}\right)\right), 
\end{equation}
where the multiplier $\e^{\ii(\lambda-\pi) d_i / 2}$ intrinsically indicates the phase of operator $\overline{\Lambda(\lambda)}$ for $i\in[l]$ since $\arg(1-\e^{\ii\lambda}) = (\lambda - \pi) /2$ for $\lambda \in [0,\pi]$. This expansion in Eq.~\eqref{eq:expansion} not only rotates the component time series, but also approximates $\overline{\Lambda(\lambda)}$ with a smaller limiting variance.

\subsection{Aperiodic Spectrum Estimation}
We proceed to calculate the spectrum density $f_X(\lambda)$. In general, the spectral density estimators comprise three categories, that is, parametric, semi-parametric, and non-parametric approaches. The parametric approaches usually employ the following periodogram estimation $\mathbf{I}_T$ (w.r.t. the discrete Fourier transform of $\mathbf{X}_t$ at frequency $\lambda_j$) to estimate $f_X$
\begin{equation} \label{eq:I}
	\mathbf{I}_T (\lambda_j) = \frac{1}{2\pi T} \sum_{t=0}^T \sum_{s=0}^T \langle \mathbf{X}_t, \mathbf{X}_s \rangle \e^{\ii (t-s) \lambda_j} .
\end{equation}
However, the periodogram estimation $\mathbf{I}_T(\lambda)$ is not a consistent estimator for $f_X(\lambda)$ despite asymptotically unbiased. To ensure the consistency and asymptotic normality of the memory parameter $\boldsymbol{d}$, one has to force an admissible spectral density $f_X(\lambda)$ of the observations and assume Gaussianity for the source process $\boldsymbol{\epsilon}_t$, detailed in Appendix~\ref{app:assumptions}. Besides, the discrete Fourier transformation can only obtain what frequency components are contained in the whole process but does not know the time when each component appears. Thereby, the conventional statistical estimators have inherent defects in processing non-stationary signals. To alleviate this issue, some researchers adapt the (semi-parametric) tapered periodogram estimation for $f_X(\lambda)$ as follows
\begin{equation} \label{eq:J'}
	\mathbf{I}_T'(\lambda_j) = \frac{1}{2\pi T} \sum_{t=0}^T \sum_{s=0}^T \langle \boldsymbol{q}_t \odot \mathbf{X}_t, \boldsymbol{q}_s \odot \mathbf{X}_s \rangle \e^{\ii (t-s) \lambda_j} ,
\end{equation}
where $\odot$ denotes the Hadamard product,
\[
\boldsymbol{q}_t(\lambda) = \left( \frac{\hbar_i(t)}{\sqrt{\sum_{s=1}^{t} \hbar_i(s)^{2}}} \right)_{i=1}^l, \quad\text{and}\quad \hbar_i(t) = \left\{ \begin{aligned}
	\left[ 1 - \cos\left( 2\pi t /T \right) \right] /2, &~ t \leq T/2; \\
	\hbar_i(1-t/T) \quad\quad, &~ t > T/2 . 
\end{aligned}\right.
\]
Here, the function $\hbar_i(t)$ indicates the Cosine-Hanning function~\citep{hurvich1995:cosine}, which works by reducing the bias of the periodogram function in Eq.~\eqref{eq:I} via a spectral window. Since the bias reduces along with an augmentation of the variance~\citep{velasco1999:cosine}, the tapered periodogram led by Eq.~\eqref{eq:J'} is still an inconsistent estimator of the spectral density $f_X$, leading to the pendent entanglement of the Gaussianity assumption.

To handle the non-stationary signals while achieving the statistical consistency, we present a non-parametric spectral density estimator, which has the discrete formation as follows
\begin{equation} \label{eq:J}
	\mathbf{J}_T (\lambda_j) = \sum_{(i,\kappa) \in \mathfrak{J}_T} \tau(\cdot; \alpha_{i,\kappa},\rho_{i,\kappa} ) \varphi_{i,\kappa}(\lambda_j) ,
\end{equation}
where $i$ and $\kappa$ are the scaled and shifted parameters in discrete wavelet-threshold transformation, respectively, the set $\mathfrak{J}_T$ is defined as $\mathfrak{J}_T = \{ (i,\kappa) \mid 2^i \leq C T^{1-\delta} \}$ for some $C>0$ and $\delta>0$ from~\citep{neumann1996}, the basis function $\varphi_{i,\kappa}(\lambda) = 2^{i/2} \varphi(2^i \lambda - \kappa)$ is normalized to square-integrate to one, which satisfies $\int_{-\pi}^{\pi} \varphi_{0,\kappa}(\lambda) \dif \lambda =1$ and $\int_{-\pi}^{\pi} \varphi_{i,\kappa}(\lambda) \lambda^s \dif \lambda =0$ for $i \neq 0$ and $s\in \mathbb{N}^+$, $\tau(\cdot; \alpha_{i,\kappa},\rho_{i,\kappa} )$ is an hard or soft threshold function~\citep{vidakovic1999:wavelet} w.r.t. the coefficients $\alpha_{i,\kappa}$ and threshold $\rho_{i,\kappa} \propto \sqrt{2 \log(|\mathfrak{J}_T|_{\#}) }$. The spectral density estimator $\mathbf{J}_T$ employs a collection of finite-length wavelet basis functions, described by Eq.~\eqref{eq:I}, instead of the infinite trigonometric ones in discrete Fourier transformation, and thus \textbf{good at extracting the aperiodic spectrum of the non-stationary observations}. On the other hand, it is well known that choosing appropriate thresholds $\rho_{i,\kappa}$ to smooth the periodogram is difficult because non-parametric spectral estimation suffers from problems similar to curve estimation with a highly heteroscedastic and non-Gaussian error structure. In Subsection~\ref{subsec:consistency-an} and Appendix~\ref{app:convergence}, we theoretically show that $\mathbf{J}_T$ is a consistent spectral density estimator when constructing the empirical thresholds $\widehat{\rho}_{i,\kappa}$ as the local weighted $l_1$ norms of the periodogram
\begin{equation} \label{eq:rho_text}
	\widehat{\rho}_{i,\kappa} = \Theta(T^{-1/2}) \cdot \left( \int_{-\pi}^{\pi} \varphi_{i,\kappa}(\lambda) \mathbf{I}_T(\lambda) \dif \lambda \right) \cdot \sqrt{2 \log(|\mathfrak{J}_T|_{\#}) }  .
\end{equation}

Invoking Eqs.~\eqref{eq:expansion} and~\eqref{eq:J} into Eq.~\eqref{eq:starting}, we can empirically rewrite the consensus as follows
\[
\mathbf{J}_T\left(\lambda_j\right) \approx f_X(\lambda_j) \sim \Psi_{j}(\boldsymbol{d}) ~\widehat{G}~ \overline{\Psi_{j}(\boldsymbol{d})}^{\top} ,
\]
where
\[
\Psi_{j}(\boldsymbol{d}) = \diag_{i \in\{1, \cdots, l\}} \left\{ \lambda_j^{-d_i} \e^{\ii\left(\pi-\lambda_j\right) d_i / 2}  \right\} \in \mathbb{C}^{l \times l} \approx \Lambda(\lambda_j) .
\]
In general, we can solve this issue by empirically maximizing the following Gaussian log-likelihood function localized to the origin
\begin{equation} \label{eq:LL_J}
	\begin{aligned}
		LL_m^J(G, \boldsymbol{d})
		&= 
		\frac{1}{m} \sum_{j=1}^{m} \bigg\{ \log \Big| \Lambda(\lambda_j) ~G~ \overline{\Lambda(\lambda_j)} \Big| + \tr\left[ \left(\Lambda(\lambda_j) ~G~ \mathbf{J}_T\left(\lambda_j\right) \overline{\Lambda(\lambda_j)} \right)^{-1} \right] \bigg\} \\
		&= \frac{1}{m} \sum_{j=1}^{m} \bigg\{ \log \Big| \Lambda(\lambda_j) ~G~ \overline{\Lambda(\lambda_j)} \Big| + \tr\left[ G^{-1} \re\left[ \left(\Lambda(\lambda_j) \mathbf{J}_T\left(\lambda_j\right) \overline{\Lambda(\lambda_j)} \right)^{-1} \right] \right] \bigg\} ,
	\end{aligned}
\end{equation}
where $m = |\{\lambda_j\}|_{\#}$ denotes the number of empirical frequencies $\{\lambda_j\}$. So, the estimation value $\widehat{\boldsymbol{d}}_{\mathrm{ASE}}$ of the memory parameter is the minimization of the following function
\begin{equation} \label{eq:ASE}
	\widehat{\boldsymbol{d}}_{\textrm{ASE}} = \arg\min_{\boldsymbol{d}} \left\{ \log \big| \widehat{G}_{\textrm{ASE}}(\boldsymbol{d}) \big| - \frac{2}{m} \sum_{i=1}^{l} \sum_{j=1}^{m} d_i \log \lambda_j \right\} ,	
\end{equation}
where
\[
\widehat{G}_{\textrm{ASE}}(\boldsymbol{d}) = \frac{1}{m} \sum_{j=1}^{m} \re\left[\Psi_{j}(\boldsymbol{d})^{-1} \mathbf{J}_T(\lambda_j) \overline{\Psi_{j}(\boldsymbol{d})}^{-1}\right] .
\]

We call Eq.~\eqref{eq:ASE} attributes to the \textit{Aperiodic Spectrum Estimation} (ASE) for the memory parameters and mark it by the subscript \texttt{ASE}. Algorithm~\ref{alg:ASE} displays the implementation procedure for calculating the estimation value $\widehat{\boldsymbol{d}}_{\textrm{ASE}}$. The details are complemented as follows.
\begin{itemize}
	\item \textbf{About the empirical spectral density estimation.} Appendix~\ref{app:convergence} details the consistent analysis for $\rho_{i,\kappa}$ and provides a concrete implementation $\widehat{\rho}_{i,\kappa}$ via the discrete Fourier transformation. By exploiting the fast algorithm in~\citep{coifman1995}, one can filter the Fourier periodogram $\mathbf{I}_T(\lambda_j)$ one scale at a timestamp, and thus, the computational complexity of Step 1 - Step 5 would become $\mathcal{O}(T^2 \log T + T^l \log T) )$, against the Fourier periodogram with $\mathcal{O}(T^l \log T)$ and the tapered periodgram with $\mathcal{O}(T^l (\log T)^l)$.
	\item \textbf{About the $\widehat{G}_{\textrm{ASE}}$.} The computational complexity for computing $\widehat{G}_{\textrm{ASE}}$ can be reduced to $\mathcal{O}(ml^2)$ by using the fast algorithm presented by Alman~\citep{alman2021} and sparse decomposition against $\mathcal{O}(ml^3)$ using the standard matrix multiplication algorithm.
	\item \textbf{About the $\widehat{\boldsymbol{d}}_{\textrm{ASE}}$.} Here, we employ the standard gradient descent algorithms to solve the optimization described by~Eq.~\eqref{eq:ASE}.
\end{itemize}
\begin{algorithm}[t]
	\footnotesize
	\caption{Aperiodic Semi-parametric Estimation for $\boldsymbol{d}$}
	\label{alg:ASE}
	\begin{algorithmic}
		\renewcommand{\algorithmicrequire}{\textbf{Input:}}
		\REQUIRE
		Input data $\{\mathbf{X}_t\}_{t=0}^{T}$, discrete Fourier frequency $\{\lambda_j\}_{j=1}^m$, and a collection of wavelet basis $\{ \varphi_{i,\kappa} \}$; Hyper-parameters $C$, $\delta$, $\kappa$.
		\renewcommand{\algorithmicensure}{\textbf{Output:}}
		\ENSURE
		Estimation value $\boldsymbol{d}_{\textrm{ASE}}$.
		\renewcommand{\algorithmicrequire}{\textbf{Procedure:}}
		\REQUIRE ~\\
		\STATE
		\textbf{1}: Compute the periodogram $\mathbf{I}_T(\lambda_j)$ at the Fourier frequency $\lambda_j = 2 \pi j /T$.
		\STATE
		\textbf{2}: Construct the indicator set $\mathfrak{J}_T$, where $(i,\kappa) \in \mathfrak{J}_T$.
		\STATE
		\textbf{3}: Compute the standard discrete wavelet transformation coefficient $\widehat{\alpha}_{i,\kappa}$ of $\mathbf{I}_T(\lambda_j)$ via a fast algorithm provided by~\citet{coifman1995}.
		\STATE
		\textbf{4}: Compute the threshold $\widehat{\rho}_{i,\kappa} \propto C\sqrt{2 \log(|\mathfrak{J}_T|_{\#}) }$ with $C = \mathcal{O}(T^{-1/2})$ from Eq.~\eqref{eq:rho_text} and Theorem~\ref{thm:convergence}.
		\STATE
		\textbf{5}: Compute the empirical threshold function $\tau(\cdot; \widehat{\alpha}_{i,\kappa},\widehat{\rho}_{i,\kappa} )$ via hard or soft threshold rules. 
		\STATE
		\textbf{6}: Compute $\widehat{G}_{\textrm{ASE}}$ according to Eq.~\eqref{eq:ASE}.
		\STATE
		\textbf{7}: Compute $\widehat{\boldsymbol{d}}_{\textrm{ASE}}$ by solving the minimization optimization described in Eq.~\eqref{eq:ASE}.
	\end{algorithmic}
\end{algorithm}

\subsection{Generalized ARISE Models}
Provided the parametric integrated process in Eq.~\eqref{eq:starting} and the aperiodic spectrum estimation in Eq.~\eqref{eq:ASE}, we can generate the \textit{ApeRIodic SEmi-parametric} (ARISE) process. As mentioned above, our proposed ARISE process is capable of modeling the sequential data with long-term memory, non-stationarity, and aperiodic spectrum, marked in bold, leading to a powerful tool for investigating the efficient market hypothesis. 

Next, we proceed to develop some generalized formations of the ARISE process. The first one is the ARISE-ARMA$(p,\boldsymbol{d},q)$ model
\[
\Lambda(\lambda) \mathbf{X}_t = \boldsymbol{\zeta}_t
\quad\text{and}\quad
\phi(\mathfrak{B}) \boldsymbol{\epsilon}_t = \boldsymbol{\epsilon}_t \psi(\mathfrak{B}) ,
\]
where $\boldsymbol{\zeta}_t$ obeys the multivariate ARMA$(p,q)$ process provided $\boldsymbol{\epsilon}_t \in \mathcal{N}(\boldsymbol{\mu}=\boldsymbol{0}, \Sigma_{\epsilon} = \mathbf{E}_{l \times l})$ for the $l$-dimensional unit matrix $\mathbf{E}_{l \times l}$ and $t \in \mathbb{N}^+$, 
\[
\phi(\mathfrak{B}) = \left( \mathbf{W}_0 + \sum_{k=1}^{p} \mathbf{W}_k \mathfrak{B}^k \right)
\quad\text{and}\quad
\psi(\mathfrak{B}) = \left( \mathbf{V}_0 + \sum_{k=1}^{q} \mathbf{V}_k \mathfrak{B}^k \right)
\]
are the corresponding matrix polynomials with lag hyper-parameters $p,q \in \mathbb{N}^+$. Under the standard stationarity and invertibility conditions~\citep{box2015} on the matrix polynomials $\phi(\mathfrak{B})$ and $\psi(\mathfrak{B})$, respectively, the process $\{ \mathbf{X}_t \}$ can be represented as
\begin{equation} \label{eq:MARFIMA}
	\mathbf{X}_t = \diag_{i \in\{1, \cdots, l\}} \left\{ (1 - \mathfrak{B})^{-d_i} \right\} \phi^{-1}(\mathfrak{B}) ~\boldsymbol{\epsilon}_t~ \psi(\mathfrak{B}) .
\end{equation}
Along to the line of this thought, we heuristically generalize the ARISE process with the machine learning model $\Theta$ as follows
\begin{equation} \label{eq:theta}
	\diag_{i \in\{1, \cdots, l\}} \left\{ (1 - \mathfrak{B})^{d_i} \right\} \mathbf{X}_t = \Theta( \boldsymbol{\epsilon}_t ) .
\end{equation}	
Here, the model $\Theta$ works like a transformer, which transforms Gaussian white noise $\{\boldsymbol{\epsilon}_t\}$ whose spectral density is constant to match the behavior of $\{\mathbf{X}_t\}_{t=0}^{\infty}$. The key difference between Eqs.~\eqref{eq:MARFIMA} and~\eqref{eq:theta} is that the model $\Theta$ is not constrained to work with a linear representation of the data. We call Eq.~\eqref{eq:theta} the ARISE-$\Theta$ model, e.g., ARISE-RNN and ARISE-LSTM. In the following, we will utilize the generalized ARISE models to explore the real-world applications of the ARISE process.

\section{Theoretical Analysis}  \label{sec:TA}
Here, we are going to theoretically show that our proposed ARISE process is a well-posed model. This section consists of three parts. Subsection~\ref{subsec:convergence} shows the consistency of $\mathbf{J}_T$ and the mean-square convergence of $\widehat{G}_{\textrm{ASE}}$. Subsection~\ref{subsec:consistency-an} demonstrates the consistency and asymptotic normality of $\widehat{\boldsymbol{d}}_{\textrm{ASE}}$. Subsection~\ref{subsec:MC} conducts the common-used Monte-Carlo study to evaluate the finite-sample performance and robustness of $\widehat{\boldsymbol{d}}_{\textrm{ASE}}$ against heavy-tailed marginals. For formal, let $G^0$ and $\boldsymbol{d}^0 = (d_1^0, \dots, d_l^0)^{\top}$ denote the true parameters corresponding to $\widehat{G}_{\mathrm{ASE}}$ and $\widehat{\boldsymbol{d}}_{\mathrm{ASE}}$, respectively.

\subsection{Mean-square Convergence} \label{subsec:convergence}
We start our analysis with some valid assumptions as follows.
\begin{assumption} \label{chara:1}
	For $k\in\mathbb{N}^+$, we assume that 
	\[
	\sup_{1\leq t_1 < \infty} \left( \sum_{t_2,\dots,t_k =1} \Big| \mathrm{CUM}(\mathbf{X}_{t_1}, \dots, \mathbf{X}_{t_k} ) \Big| \right) \leq C^k(k!)^{1+\gamma},
	\]
	where $C$ is a generic positive constant and $\gamma>0$.
\end{assumption}
Assumption~\ref{chara:1} shows that the asymptotic normality of the local cumulative sums of $\mathbf{X}_t$ is uniform over time $t$ to some extent~\citep{neumann1996}. It's observed that if $\{\mathbf{X}_t\}_{t=0}^{\infty}$ is $\alpha$-mixing with an appropriate rate and its marginal distribution is Gaussian, exponential, gamma, or inverse Gaussian, then $\gamma$ can be set equal to zero. For heavier-tailed marginals, we should employ a positive $\gamma$.

We present our first main theorem as follows.
\begin{theorem} \label{thm:convergence}
	Let $\{\mathbf{X}_t\}_{t=0}^{\infty}$ be an $l$-dimensional process specified by Eq.~\eqref{eq:starting}, which meets Assumption~\ref{chara:1}, and $f_X$ is the corresponding spectral density matrix, which satisfies that $f_X(\lambda)>0$ and $f_X(\lambda)$ is of finite total variation over $[-\pi,\pi]$. Then there exists some threshold $\rho_{i,\kappa}$ in which $\rho_{i,\kappa} \propto \sqrt{2 \log(|\mathfrak{J}_T|_{\#}) }$, such that
	\[
	\sup_{f_X \in \mathcal{B}_{p,q}^n(\mathbb{R};R)} \left\{ \mathbb{E} \left[ \left\| \widehat{G}_{\textrm{ASE}} - G^0 \right\|_{L_2([-\pi,\pi])} \right] \right\} = \mathcal{O} \left( (\log T/T)^{2n/(2n+1)} \right) ,
	\]
	where $\mathcal{B}_{p,q}^n(\mathbb{R};R)$ is a Besov space with $p,q,m \geq 1$ and a radius scalar $R>0$, detailed in Appendix~\ref{app:convergence}. Furthermore, if $\boldsymbol{d}^0 \in \Omega_{\beta}$, we have
	\[
	\widehat{G}_{\textrm{ASE}}(\boldsymbol{d}^0) = G^0 + \eta,
	\]
	where $\eta$ is an infinitesimal number that converges in probability to zero at a constant rate, denoted as $\eta=\op(1)$.
\end{theorem} 
Theorem~\ref{thm:convergence} establishes the consistency of $\widehat{G}_{\textrm{ASE}}$, including a guarantee that $\widehat{G}_{\textrm{ASE}}$ has the near-optimal rate of mean-square convergence and a consistent approximation in probability. There optimal rate of mean-square convergence, alternatively known as minimax rate is $T^{-2n/(2n+1)}$.  Appendix~\ref{app:convergence} details the complete proof of Theorem~\ref{thm:convergence}. Notice that $\widehat{G}_{\textrm{ASE}}$ is data-driven and Theorem~\ref{thm:convergence} holds without any Gaussian assumption, which sheds some insights on the robustness of $\widehat{\boldsymbol{d}}_{\textrm{ASE}}$ against heavier-tailed marginals. We demonstrate this conjecture by the typical Monte Carlo study in Subsection~\ref{subsec:MC}. Besides, this theorem constitutes a solid stone for proving the consistency and asymptotic normality of $\widehat{\boldsymbol{d}}_{\textrm{ASE}}$ as discussed in Subsection~\ref{subsec:consistency-an}.

\subsection{Consistency and Asymptotic Normality} \label{subsec:consistency-an}
In this subsection, we proceed to demonstrate the consistency of $\widehat{\boldsymbol{d}}_{\textrm{ASE}}$, described by Eq.~\eqref{eq:ASE}, and then, present a sufficient condition for its asymptotic normality, with $\mathbf{J}_T$ as an estimator of the spectral density function $f_X(\lambda)$ satisfying a single regularity condition. Limited to the space, the assumptions, which are exactly the same as those of~\citep{shimotsu2007:GSE}, are placed in Appendix~\ref{app:assumptions}.

We now present the consistency and asymptotic normality theorem for $\widehat{\boldsymbol{d}}_{\textrm{ASE}}$ as follows.
\begin{theorem} \label{thm:consistency-an}
	Let Assumptions~\ref{ass:1}-\ref{ass:4} hold. Then we have
	\[
	\widehat{\boldsymbol{d}}_{\textrm{ASE}} \stackrel{\mathrm{P}}{\longrightarrow} \boldsymbol{d}^0 \quad\text{as}\quad T \rightarrow \infty .
	\]
	Let Assumptions~\ref{ass:3} and~\ref{ass:B1}--\ref{ass:B4} hold. We have
	\[
	\sqrt{m} \left( \widehat{\boldsymbol{d}}_{\textrm{ASE}} - \boldsymbol{d}^0 \right) \stackrel{\mathrm{d}}{\longrightarrow} \mathcal{N}\left(0, \Sigma^{-1} \right) \quad\text{and}\quad
	\widehat{G} \left( \widehat{\boldsymbol{d}}_{\textrm{ASE}} \right) \stackrel{\mathrm{P}}{\longrightarrow} G^0  \quad\text{as}\quad T \to \infty,
	\]
	where
	\[
	\Sigma = \frac{4+\pi^2}{2} G^0 \odot\left(G^0\right)^{-1} + \frac{4-\pi^{2}}{2} \mathbf{1}_{l \times l}  .
	\]
\end{theorem}
This proof ideas can be summarized as follows. For consistency, it is equivalent to show that $\mathbb{P} ( \| \widehat{\boldsymbol{d}}_{\mathrm{ASE}} - \boldsymbol{d}_{0} \|_{\infty} > \delta ) \to 0$ as $T \to \infty$. Observe that
\[
\resizebox{1\hsize}{!}{$
	\begin{aligned}
		&\mathbb{P} \left(\left\| \widehat{\boldsymbol{d}}_{\mathrm{ASE}} - \boldsymbol{d}_{0} \right\|_{\infty} > \delta\right) \\
		&~\leq \mathbb{P} \left\{ \inf_{\overline{\Omega_{\delta}} \cap \Omega_{\beta}} \left\{ \left( \log \big| \widehat{G}_\textrm{ASE}(\boldsymbol{d}) \big| - \frac{2}{m} \sum_{i=1}^{l} \sum_{j=1}^{m} d_i \log \lambda_j \right) - \left( \log \big| \widehat{G}_\textrm{ASE}(\boldsymbol{d}^0) \big| - \frac{2}{m} \sum_{i=1}^{l} \sum_{j=1}^{m} d_i^0 \log \lambda_j \right) \right\} \leq 0 \right\} ,
	\end{aligned} $}
\]
where $\Omega_{\delta} = \{ \boldsymbol{d} \mid \| \boldsymbol{d} - \boldsymbol{d}^0 \|_{\infty} > \delta \}$.
Thus, it suffices to show that
\[
\log \big| \widehat{G}_\textrm{ASE}(\boldsymbol{d}) \big| - \log \big| \widehat{G}_\textrm{ASE}(\boldsymbol{d}^0) \big| = 0 + \op(1)
\quad\text{and}\quad
\frac{2}{m} \sum_{i=1}^{l} \sum_{j=1}^{m} (d_i^0 - d_i) \log \lambda_j = 0 + \op(1).  
\]
From Theorem~\ref{thm:convergence}, we have proved that $J_T(\lambda) = f_X(\lambda) + \op(T^{-\beta})$ and $\widehat{G}_{\textrm{ASE}}(\boldsymbol{d}^0) = G^0 + \op(1)$, for $\beta \in (0,1)$ and $\boldsymbol{d}^0 \in \Omega_{\beta}$, and thus, it is easily to demonstrate the consistency of $\widehat{\boldsymbol{d}}_{\textrm{ASE}}$. Further, for some $\boldsymbol{d}$ such that $\| \boldsymbol{d} - \boldsymbol{d}^0 \|_{\infty} \leq \| \widehat{\boldsymbol{d}}_{\textrm{ASE}} -\boldsymbol{d}^0 \|_{\infty}$, with probability tending to one, one has
\[
\begin{aligned}
	0 &= \left.\frac{\dif }{\dif \boldsymbol{d} }\right|_{\widehat{\boldsymbol{d}}_{\textrm{ASE}}} \left\{ \log \big| \widehat{G}_\textrm{ASE}(\boldsymbol{d}) \big| - \frac{2}{m} \sum_{i=1}^{l} \sum_{j=1}^{m} d_i \log \lambda_j \right\} \\
	&= \Bigg[ \left.\frac{\dif }{\dif \boldsymbol{d}}\right|_{\boldsymbol{d}^0} + \left( \widehat{\boldsymbol{d}}_{\mathrm{ASE}} - \boldsymbol{d}^0 \right) \left.\frac{\dif^{2} }{\dif \boldsymbol{d}^{\top} \dif \boldsymbol{d} }\right|_{\bar{\boldsymbol{d}}} \Bigg] \left( \log \big| \widehat{G}_\textrm{ASE}(\boldsymbol{d}) \big| - \frac{2}{m} \sum_{i=1}^{l} \sum_{j=1}^{m} d_i \log \lambda_j \right) ,
\end{aligned}
\]
as $T$ goes to infinity. For asymptotic normality, it is observed that $\widehat{\boldsymbol{d}}_{\textrm{ASE}}$ has the stated limiting distribution if the followings hold
\[
\left.\sqrt{m}~ \frac{\partial }{\partial \boldsymbol{d}}\right|_{\boldsymbol{d}^0} \left( \log \big| \widehat{G}_\textrm{ASE}(\boldsymbol{d}) \big| - \frac{2}{m} \sum_{i=1}^{l} \sum_{j=1}^{m} d_i \log \lambda_j \right)  \stackrel{\mathrm{d}}{\longrightarrow} \mathcal{N}(0, \Sigma^{-1})
\]
and 
\[
\left. \frac{\partial^{2} }{\partial \boldsymbol{d}^{\top} \partial \boldsymbol{d} } \right|_{\bar{\boldsymbol{d}}} \left( \log \big| \widehat{G}_\textrm{ASE}(\boldsymbol{d}) \big| - \frac{2}{m} \sum_{i=1}^{l} \sum_{j=1}^{m} d_i \log \lambda_j \right) \stackrel{\mathrm{P}}{\longrightarrow} \Sigma^{-1} ,
\]
as $T \to \infty$. The full proof of Theorem~\ref{thm:consistency-an} is completed in Appendix~\ref{app:consistency-an}.

\subsection{Monte Carlo Study} \label{subsec:MC}
Inspired from~\citep{lo1989:MC}, we conduct the Monte Carlo study on some simulated data to evaluate the finite-sample performance and robustness of $\widehat{\boldsymbol{d}}_{\mathrm{ASE}}$ against heavy-tailed marginals. Practically, we set $C=1$ and $\delta=0.01$ in Algorithm~\ref{alg:ASE} as the default scalars for the better practical performance. The contenders we adapt are the $\widehat{\boldsymbol{d}}_{\mathrm{GSE}}$ and $\widehat{\boldsymbol{d}}_{\mathrm{TSE}}$, detailed in Appendix~\ref{app:GSE}. Here, we generate $T = 2^{10}$ observations from Eq.~\eqref{eq:starting} where the fractionally differencing parameter $\boldsymbol{d}^0$ are selected from the set $\{ (0.1, 0.3), (0.2,0.4), (0.1,0.4) \}$ and the source process is bivariate Gaussian but the marginals are not. The marginal candidates include the Student's $t$ distribution with 3 and 7 degrees of freedom (i.e., $t_3$ and $t_7$, respectively), the Standard Logistic distribution (i.e., $sl_{(0,1)}$) with density $\mathrm{sech}(x/2)^2/4$, and the hyperbolic-secant distribution (i.e., $hs_1$) with density $\mathrm{sech}(\pi x/2)/2$, for $x \in \mathbb{R}$. These candidate marginals are heavier-tailed than the Gaussian distribution, with excess of kurtosis $\infty$ for $t_3$, 2 for $t_7$ and $hs_1$, and $1.2$ for $sl_{(0,1)}$. The source process is generated by coupling the marginals above with a Copulas-based approach from~\citep{lopes2013:copulas}, and its covariance is selected from the set $\tau \in \{0.2, 0.4, 0.6\}$. Each experiment is performed 500 trails.

\begin{table}[!htb]
	\centering
	\caption{Performance of the different Estimators against heavy-tailed marginals.}
	\label{tab:MC}
	\resizebox{\textwidth}{!}{%
		\begin{tabular}{@{}ccccccc@{}}
			\toprule
			$\tau$ & $\boldsymbol{d}^0$ & $\widehat{\boldsymbol{d}}$ & $t_3$ & $t_7$ & $sl_{(0,1)}$ & $hs_1$ \\
			\midrule
			\multicolumn{1}{c|}{\multirow{9}{*}{$\tau=0.2$}} & \multicolumn{1}{c|}{\multirow{3}{*}{(0.1,0.3)}} & \multicolumn{1}{c|}{$\widehat{\boldsymbol{d}}_{\mathrm{GSE}}$} & $(0.0962 \pm 0.0347, 0.2838 \pm 0.0270)$
			& $(0.0959 \pm 0.0367, 0.2841 \pm 0.0388)$
			& $(0.0958 \pm 0.0375, 0.2838 \pm 0.0385)$
			& $(0.0959 \pm 0.0385, 0.2841 \pm 0.0303)$ \\
			\multicolumn{1}{c|}{} & \multicolumn{1}{c|}{} & \multicolumn{1}{c|}{$\widehat{\boldsymbol{d}}_{\mathrm{TSE}}$} & $(0.1008 \pm 0.0376, 0.2919 \pm 0.0385)$
			& $(0.0938 \pm 0.0382, 0.2849 \pm 0.0380)$
			& $(0.0937 \pm 0.0367, 0.2883 \pm 0.0372)$
			& $(0.0945 \pm 0.0386, 0.2857 \pm 0.0382)$ \\
			\multicolumn{1}{c|}{} & \multicolumn{1}{c|}{} & \multicolumn{1}{c|}{$\widehat{\boldsymbol{d}}_{\mathrm{ASE}}$} & $(0.0973 \pm 0.0270, 0.2928 \pm 0.0292)$
			& $(0.0971 \pm 0.0267, 0.2930 \pm 0.0286)$
			& $(0.0971 \pm 0.0299, 0.2931 \pm 0.0270)$
			& $(0.0971 \pm 0.0270, 0.2928 \pm 0.0282)$  \\
			\cmidrule(l){2-7} 
			\multicolumn{1}{c|}{} & \multicolumn{1}{c|}{\multirow{3}{*}{(0.2,0.4)}} & \multicolumn{1}{c|}{$\widehat{\boldsymbol{d}}_{\mathrm{GSE}}$} & $(0.1917 \pm 0.0271, 0.3801 \pm 0.0293)$
			& $(0.1913 \pm 0.0273, 0.3804 \pm 0.0291)$
			& $(0.1913 \pm 0.0268, 0.3804 \pm 0.0270)$
			& $(0.1913 \pm 0.0271, 0.3803 \pm 0.0272)$  \\
			\multicolumn{1}{c|}{} & \multicolumn{1}{c|}{} & \multicolumn{1}{c|}{$\widehat{\boldsymbol{d}}_{\mathrm{TSE}}$} & $(0.1932 \pm 0.0373, 0.3900 \pm 0.0379)$
			& $(0.1909 \pm 0.0376, 0.3869 \pm 0.0385)$
			& $(0.1925 \pm 0.0371, 0.3873 \pm 0.0382)$
			& $(0.1921 \pm 0.0375, 0.3907 \pm 0.0377)$  \\
			\multicolumn{1}{c|}{} & \multicolumn{1}{c|}{} & \multicolumn{1}{c|}{$\widehat{\boldsymbol{d}}_{\mathrm{ASE}}$} & $(0.1945 \pm 0.0270, 0.3969 \pm 0.0288)$
			& $(0.1944 \pm 0.0268, 0.3985 \pm 0.0288)$
			& $(0.1944 \pm 0.0272, 0.3986 \pm 0.0268)$
			& $(0.1944 \pm 0.0263, 0.3983 \pm 0.0277)$  \\
			\cmidrule(l){2-7} 
			\multicolumn{1}{c|}{} & \multicolumn{1}{c|}{\multirow{3}{*}{(0.1,0.4)}} & \multicolumn{1}{c|}{$\widehat{\boldsymbol{d}}_{\mathrm{GSE}}$} & $(0.0965 \pm 0.0275, 0.3804 \pm 0.0289)$
			& $(0.0962 \pm 0.0270, 0.3806 \pm 0.0291)$
			& $(0.0962 \pm 0.0271, 0.3807 \pm 0.0292)$
			& $(0.0962 \pm 0.0270, 0.3806 \pm 0.0289)$  \\
			\multicolumn{1}{c|}{} & \multicolumn{1}{c|}{} & \multicolumn{1}{c|}{$\widehat{\boldsymbol{d}}_{\mathrm{TSE}}$} & $(0.1011 \pm 0.0371, 0.3939 \pm 0.0386)$
			& $(0.0907 \pm 0.0376, 0.3853 \pm 0.0384)$
			& $(0.0961 \pm 0.0370, 0.3912 \pm 0.0380)$
			& $(0.0925 \pm 0.0369, 0.3885 \pm 0.0382)$ \\
			\multicolumn{1}{c|}{} & \multicolumn{1}{c|}{} & \multicolumn{1}{c|}{$\widehat{\boldsymbol{d}}_{\mathrm{ASE}}$} & $(0.0975 \pm 0.0268, 0.3972 \pm 0.0301)$
			& $(0.0974 \pm 0.0270, 0.3989 \pm 0.0303)$
			& $(0.0974 \pm 0.0270, 0.3991 \pm 0.0292)$
			& $(0.0974 \pm 0.0267, 0.3987 \pm 0.0295)$ \\
			\midrule
			\multicolumn{1}{c|}{\multirow{9}{*}{$\tau=0.4$}} & \multicolumn{1}{c|}{\multirow{3}{*}{(0.1,0.3)}} & \multicolumn{1}{c|}{$\widehat{\boldsymbol{d}}_{\mathrm{GSE}}$} & $(0.0969 \pm 0.0257, 0.2841 \pm 0.0274)$
			& $(0.0967 \pm 0.0252, 0.2848 \pm 0.0271)$
			& $(0.0967 \pm 0.0253, 0.2849 \pm 0.0272)$
			& $(0.0967 \pm 0.0252, 0.2848 \pm 0.0270)$  \\
			\multicolumn{1}{c|}{} & \multicolumn{1}{c|}{} & \multicolumn{1}{c|}{$\widehat{\boldsymbol{d}}_{\mathrm{TSE}}$} & $(0.0919 \pm 0.0356, 0.2868 \pm 0.0362)$
			& $(0.0936 \pm 0.0354, 0.2857 \pm 0.0359)$
			& $(0.0920 \pm 0.0351, 0.2847 \pm 0.0359)$
			& $(0.0998 \pm 0.0352, 0.2935 \pm 0.0359)$  \\
			\multicolumn{1}{c|}{} & \multicolumn{1}{c|}{} & \multicolumn{1}{c|}{$\widehat{\boldsymbol{d}}_{\mathrm{ASE}}$} & $(0.0979 \pm 0.0250, 0.2929 \pm 0.0273)$
			& $(0.0979 \pm 0.0248, 0.2946 \pm 0.0269)$
			& $(0.0979 \pm 0.0247, 0.2944 \pm 0.0268)$
			& $(0.0979 \pm 0.0247, 0.2948 \pm 0.0268)$  \\
			\cmidrule(l){2-7} 
			\multicolumn{1}{c|}{} & \multicolumn{1}{c|}{\multirow{3}{*}{(0.2,0.4)}} & \multicolumn{1}{c|}{$\widehat{\boldsymbol{d}}_{\mathrm{GSE}}$} & $(0.1924 \pm 0.0252, 0.3804 \pm 0.0273)$
			& $(0.1922 \pm 0.0250, 0.3811 \pm 0.0262)$
			& $(0.1922 \pm 0.0271, 0.3811 \pm 0.0270)$
			& $(0.1922 \pm 0.0268, 0.3810 \pm 0.0262)$  \\
			\multicolumn{1}{c|}{} & \multicolumn{1}{c|}{} & \multicolumn{1}{c|}{$\widehat{\boldsymbol{d}}_{\mathrm{TSE}}$} & $(0.1916 \pm 0.0353, 0.3895 \pm 0.0359)$
			& $(0.1936 \pm 0.0386, 0.3887 \pm 0.0327)$
			& $(0.1935 \pm 0.0351, 0.3892 \pm 0.0352)$
			& $(0.1919 \pm 0.0353, 0.3911 \pm 0.0324)$  \\
			\multicolumn{1}{c|}{} & \multicolumn{1}{c|}{} & \multicolumn{1}{c|}{$\widehat{\boldsymbol{d}}_{\mathrm{ASE}}$} & $(0.1952 \pm 0.0250, 0.3985 \pm 0.0257)$
			& $(0.1953 \pm 0.0251, 0.4006 \pm 0.0252)$
			& $(0.1953 \pm 0.0253, 0.4008 \pm 0.0263)$
			& $(0.1952 \pm 0.0251, 0.4004 \pm 0.0252)$  \\
			\cmidrule(l){2-7} 
			\multicolumn{1}{c|}{} & \multicolumn{1}{c|}{\multirow{3}{*}{(0.1,0.4)}} & \multicolumn{1}{c|}{$\widehat{\boldsymbol{d}}_{\mathrm{GSE}}$} & $(0.0979 \pm 0.0255, 0.3814 \pm 0.0279)$
			& $(0.0981 \pm 0.0254, 0.3823 \pm 0.0271)$
			& $(0.0981 \pm 0.0257, 0.3824 \pm 0.0282)$
			& $(0.0981 \pm 0.0254, 0.3822 \pm 0.0283)$  \\
			\multicolumn{1}{c|}{} & \multicolumn{1}{c|}{} & \multicolumn{1}{c|}{$\widehat{\boldsymbol{d}}_{\mathrm{TSE}}$} & $(0.0979 \pm 0.0361, 0.3924 \pm 0.0366)$
			& $(0.0970 \pm 0.0362, 0.3841 \pm 0.0360)$
			& $(0.0972 \pm 0.0350, 0.3918 \pm 0.0358)$
			& $(0.0974 \pm 0.0354, 0.3911 \pm 0.0352)$ \\
			\multicolumn{1}{c|}{} & \multicolumn{1}{c|}{} & \multicolumn{1}{c|}{$\widehat{\boldsymbol{d}}_{\mathrm{ASE}}$} & $(0.0991 \pm 0.0238, 0.3997 \pm 0.0251)$
			& $(0.0995 \pm 0.0230, 0.4022 \pm 0.0243)$
			& $(0.0996 \pm 0.0234, 0.4024 \pm 0.0247)$
			& $(0.0995 \pm 0.0235, 0.4019 \pm 0.0244)$ \\
			\midrule
			\multicolumn{1}{c|}{\multirow{9}{*}{$\tau=0.6$}} & \multicolumn{1}{c|}{\multirow{3}{*}{(0.1,0.3)}} & \multicolumn{1}{c|}{$\widehat{\boldsymbol{d}}_{\mathrm{GSE}}$} & $(0.1018 \pm 0.0223, 0.2878 \pm 0.0246)$
			& $(0.1029 \pm 0.0221, 0.2895 \pm 0.0245)$
			& $(0.1030 \pm 0.0223, 0.2896 \pm 0.0245)$
			& $(0.1028 \pm 0.0223, 0.2893 \pm 0.0247)$  \\
			\multicolumn{1}{c|}{} & \multicolumn{1}{c|}{} & \multicolumn{1}{c|}{$\widehat{\boldsymbol{d}}_{\mathrm{TSE}}$} & $(0.0934 \pm 0.0315, 0.2863 \pm 0.0325)$
			& $(0.0923 \pm 0.0310, 0.2895 \pm 0.0321)$
			& $(0.0947 \pm 0.0309, 0.2896 \pm 0.0321)$
			& $(0.0967 \pm 0.0309, 0.2893 \pm 0.0323)$  \\
			\multicolumn{1}{c|}{} & \multicolumn{1}{c|}{} & \multicolumn{1}{c|}{$\widehat{\boldsymbol{d}}_{\mathrm{ASE}}$} & $(0.1008 \pm 0.0223, 0.2979 \pm 0.0241)$
			& $(0.1007 \pm 0.0217, 0.3007 \pm 0.0237)$
			& $(0.1008 \pm 0.0217, 0.3009 \pm 0.0239)$
			& $(0.1007 \pm 0.0219, 0.3004 \pm 0.0238)$  \\
			\cmidrule(l){2-7} 
			\multicolumn{1}{c|}{} & \multicolumn{1}{c|}{\multirow{3}{*}{(0.2,0.4)}} & \multicolumn{1}{c|}{$\widehat{\boldsymbol{d}}_{\mathrm{GSE}}$} & $(0.1955 \pm 0.0221, 0.3842 \pm 0.0240)$
			& $(0.1963 \pm 0.0220, 0.3857 \pm 0.0241)$
			& $(0.1963 \pm 0.0222, 0.3858 \pm 0.0238)$
			& $(0.1962 \pm 0.0221, 0.3856 \pm 0.0241)$  \\
			\multicolumn{1}{c|}{} & \multicolumn{1}{c|}{} & \multicolumn{1}{c|}{$\widehat{\boldsymbol{d}}_{\mathrm{TSE}}$} & $(0.1889 \pm 0.0309, 0.3878 \pm 0.0317)$
			& $(0.1883 \pm 0.0310, 0.3883 \pm 0.0316$
			& $(0.1917 \pm 0.0311, 0.3898 \pm 0.0321)$
			& $(0.1921 \pm 0.0309, 0.3897 \pm 0.0319)$  \\
			\multicolumn{1}{c|}{} & \multicolumn{1}{c|}{} & \multicolumn{1}{c|}{$\widehat{\boldsymbol{d}}_{\mathrm{ASE}}$} & $(0.1996 \pm 0.0220, 0.4022 \pm 0.0238)$
			& $(0.2010 \pm 0.0219, 0.4025 \pm 0.0234)$
			& $(0.2011 \pm 0.0220, 0.4027 \pm 0.0238)$
			& $(0.2009 \pm 0.0219, 0.4024 \pm 0.0235)$  \\
			\cmidrule(l){2-7} 
			\multicolumn{1}{c|}{} & \multicolumn{1}{c|}{\multirow{3}{*}{(0.1,0.4)}} & \multicolumn{1}{c|}{$\widehat{\boldsymbol{d}}_{\mathrm{GSE}}$} & $(0.1050 \pm 0.0231, 0.3889 \pm 0.0279)$
			& $(0.1069 \pm 0.0233, 0.3916 \pm 0.0280)$
			& $(0.1070 \pm 0.0233, 0.3918 \pm 0.0277)$
			& $(0.1068 \pm 0.0232, 0.3914 \pm 0.0279)$  \\
			\multicolumn{1}{c|}{} & \multicolumn{1}{c|}{} & \multicolumn{1}{c|}{$\widehat{\boldsymbol{d}}_{\mathrm{TSE}}$} & $(0.0938 \pm 0.0327, 0.3890 \pm 0.0339)$
			& $(0.0988 \pm 0.0325, 0.3907 \pm 0.0338)$
			& $(0.0930 \pm 0.0326, 0.3855 \pm 0.0334)$
			& $(0.1007 \pm 0.0326, 0.3940 \pm 0.0337)$  \\
			\multicolumn{1}{c|}{} & \multicolumn{1}{c|}{} & \multicolumn{1}{c|}{$\widehat{\boldsymbol{d}}_{\mathrm{ASE}}$} & $(0.1041 \pm 0.0229, 0.4073 \pm 0.0256)$
			& $(0.1047 \pm 0.0228, 0.4047 \pm 0.0251)$
			& $(0.1048 \pm 0.0227, 0.4123 \pm 0.0254)$
			& $(0.1045 \pm 0.0228, 0.4036 \pm 0.0254)$  \\
			\bottomrule
	\end{tabular} }
\end{table}
Table~\ref{tab:MC} lists the experimental results, i.e. mean $\pm$ std of the estimators, where the smaller bias and variance imply a high performance (i.e., robustness and accuracy). It's no doubt that all estimators achieved good performance, whereas the proposed estimation value $\widehat{\boldsymbol{d}}_{\mathrm{ASE}}$ performs best against other competing approaches, achieving the smallest bias and variance (i.e., mean square error) when $\tau = 0.4$ and $0.6$. It is a laudable result for the estimation of $\boldsymbol{d}$ against heavier-tailed marginals, although in the most cases, the difference among three estimators is only on the third decimal place.

\section{Applications} \label{sec:applications}
This section provides three applications for the ARISE process, including the identification of efficient markets and two machine-learning scenarios, that is, the investigation of long-term memorability of various machine-learning models and the latent state-space model for inference and forecasting.

\subsection{Investigation of Efficient Markets} \label{subsec:EMH}
In this subsection, we apply the proposed ARISE process to investigate the efficient market hypothesis. As mentioned in Section~\ref{sec:ARISE}, the estimation value of $\boldsymbol{d}$ explicitly indicates the long-term memory of the ARISE process. Inspired by this recognition and the ideas in~\citep{greaves2019:SG}, we design the following testable criteria.
\begin{itemize}
	\item \textbf{Step 1.} Define the \textit{averaged memory statistic}
	\[
	\bar{d} = \mathrm{AVERAGE}(\widehat{\boldsymbol{d}}_{\textrm{ASE}}) = \mathbf{1}^{\top} \widehat{\boldsymbol{d}}_{\textrm{ASE}} / l ,
	\]
	where $\mathbf{1}$ is an abbreviation of the vector $\mathbf{1}_{l \times 1}$. Since $\bar{d}$ is a simple linear combination of $\widehat{\boldsymbol{d}}_{\textrm{ASE}}$, its consistency and asymptotic normality can be established by a simple argument from Theorem~\ref{thm:consistency-an}, that is,
	\[
	\bar{d} \stackrel{\mathrm{P}}{\longrightarrow} \bar{d}^0 \quad\text{and}\quad \sqrt{m} ( \bar{d} - \bar{d}^0 ) \stackrel{\mathrm{d}}{\longrightarrow} \mathcal{N}\left(0, \mathbf{1}^{\top} \Sigma^{-1} \mathbf{1} /~ l^2 \right) ,
	\]
	as $T \rightarrow \infty$, where $\bar{d}^0$ denotes the true averaged memory.
	\item \textbf{Step 2.} Provided the price data $\{ \mathbf{X}_t \}_{t=0}^{T}$ from some real-world asset markets, we mimic the concerned data by the ARISE process at time interval $[0,T]$ and calculate the averaged memory statistic $\bar{d}$. We test the null hypothesis $\mathcal{H}_0: \bar{d}=0$ the one-side alternative of long memory $\mathcal{H}_1: \bar{d}>0$ with 0.05 test level. The result ``reject $\mathcal{H}_0$" reveals that $\{ \mathbf{X}_t \}_{t=0}^{T}$ belongs to a long-term memory process, and thus, the efficient market hypothesis holds on the concerned data $\{ \mathbf{X}_t \}_{t=0}^{T}$. Otherwise, the evaluated market is non-efficient in a statistical sense.
\end{itemize}

\begin{table}[t]
	\centering
	\caption{Dependency estimation of the conducted data.}
	\label{tab:datasets}	
	\begin{tabular}{@{}cccc@{}}
		\toprule[2pt]
		Data &  Statistic $\bar{d}^*$ & $p$-value  & Market Efficiency \\
		\midrule[1.5pt]
		\texttt{Gaussian white noise}  & 0      & $<1 \times 10^{-16}$  & No  \\
		\texttt{CSI 300 Index} & 0.142 & $<1 \times 10^{-16}$  & Yes  \\
		\texttt{Winton Stock Exchange}  & 0.189    & $<1 \times 10^{-16}$  & Yes  \\
		\texttt{SSEC}  & 0.183    & $<1 \times 10^{-16}$  & Yes  \\
		\texttt{Penn TreeBank} & 0.163  & $<1 \times 10^{-16}$  & Yes  \\
		\bottomrule[2pt]
	\end{tabular}
\end{table}	

The data sets we use comprise: (1) \texttt{Gaussian White Noise} is a univariate standard Gaussian white noise sequence with length $2^{16}$. (2) \texttt{CSI 300 Index} indicates the minute-level prices of the CSI 300 Index (399300.SZ) from April 25th, 2005 to May 20th, 2014~\citep{zhang2020:hrp}. (3) \texttt{Winton Stock Exchange}, provided by the \emph{Winton Stock Market Challenge}\footnote{\url{https://www.kaggle.com/c/the-winton-stock-market-challenge}}, records the minute-level stock price sequence. (4) \texttt{Shanghai Securities Composite Index} (or equally, SSEC) that accesses from Tushare\footnote{\url{http://tushare.org/index.html}}, contains the daily log return of SSEC (000001.SH) around 150 days. (5) \texttt{Penn TreeBank} corpus consists of millions of English language text data~\citep{marcus1993}. We alter the testable criteria designed above, and perform 500 trials to calculate the statistic $\bar{d}$. Table~\ref{tab:datasets} lists the numerical statistics and hypothesis testing results, from which all real-world data sets appear the market efficiency, that is, long-term memory.

\begin{table}[t]
	\centering
	\caption{Residual estimation of the concerned models for pursuing long-range dependency.}
	\label{tab:applications}
	\resizebox{1\textwidth}{!}{	
		\begin{tabular}{@{}ccccc@{}}
			\toprule[2pt]
			Input Signals & Models &  Statistic ($\bar{d}-\bar{d}^*$) $\times 10^{-5}$ & $p$-value  & Reject $\mathcal{H}_0$ \\ \midrule[1.5pt]
			\multirow{5}*{\texttt{Gaussian White Noise}}
			& RNN   & $-34.55 \pm 460$ & 0.5353 & No  \\
			& MGU   & $-1.548 \pm 250$ & 0.6327 & No  \\
			& GRU   & $-62.05 \pm 180$ & 0.5584 & No  \\
			& LSTM  & $-1.455 \pm 310$ & 0.5412 & No  \\
			& FTNet & $-71.84 \pm 630$ & 0.5551 & No  \\
			\midrule
			\multirow{5}*{\texttt{CSI 300 Index}}
			& RNN   & $-66.47 \pm 42$ & $ 4.097 \times 10^{-2}$  & Yes \\
			& MGU   & $-8.707 \pm 4.3$ & $<1 \times 10^{-16}$ 
			& Yes  \\
			& GRU   & $-8.203 \pm 2.6$ & $<1 \times 10^{-16}$ 
			& Yes  \\
			& LSTM  & $-3.842 \pm 2.2$ & $<1 \times 10^{-16}$ 
			& Yes \\
			& FTNet & $-9.504 \pm 6.5$ & $ 4.172 \times 10^{-2}$ & Yes  \\ 
			\midrule
			\multirow{5}*{\texttt{Winton Stock Exchange}}
			& RNN   & $-46.59 \pm 34$ & $ 3.354 \times 10^{-2}$  & Yes  \\
			& MGU   & $-4.005 \pm 0.50$ & $<1 \times 10^{-16}$
			& Yes  \\
			& GRU   & $-1.837 \pm 0.12$ & $<1 \times 10^{-16}$
			& Yes  \\
			& LSTM  & $-1.968 \pm 0.10$ & $<1 \times 10^{-16}$
			& Yes \\
			& FTNet & $-9.359 \pm 0.71$ & $3.147 \times 10^{-2}$ & Yes  \\ 
			\midrule
			\multirow{5}*{\texttt{SSEC}}
			& RNN   & $-37.21 \pm 26$ & $ 3.265 \times 10^{-2}$  & Yes  \\
			& MGU   & $-5.170 \pm 0.53$ & $<1 \times 10^{-16}$
			& Yes  \\
			& GRU   & $-1.645 \pm 0.13$ & $<1 \times 10^{-16}$
			& Yes  \\
			& LSTM  & $-1.798 \pm 0.09$ & $<1 \times 10^{-16}$
			& Yes \\
			& FTNet & $-8.934 \pm 0.49$ & $2.743 \times 10^{-2}$ & Yes  \\ 
			\midrule
			\multirow{5}*{\texttt{Penn TreeBank}}
			& RNN   & $-90.19 \pm 54$ & $2.701 \times 10^{-2}$ 
			& Yes  \\
			& MGU   & $-2.358 \pm 0.82$ & $<1 \times 10^{-16}$ 
			& Yes  \\
			& GRU   & $-1.101 \pm 0.53$ & $<1 \times 10^{-16}$ 
			& Yes  \\
			& LSTM  & $-1.394 \pm 0.61$ & $<1 \times 10^{-16}$ 
			& Yes  \\
			& FTNet & $-7.388 \pm 1.40$ & $3.152 \times 10^{-2}$ & Yes  \\ 
			\bottomrule[2pt]
	\end{tabular} }
\end{table}
\subsection{Memorability of Machine Learning Models} \label{subsec:memory}
Notice that the estimation value of $\boldsymbol{d}$ not only indicates the long-term memory of the ARISE process, but also provides a means to evaluate whether the process transformed from $\{ \boldsymbol{\epsilon}_t\}$ are agnostic to the behavior of the concerned one $\{\mathbf{X}_t\}_{t=0}^{\infty}$ at relatively high frequencies~\citep{greaves2019:SG}. If the model $\Theta$ admits a temporal representation in terms of the ARISE process with explicit memory parameters, then this can be investigated by statistical tests. So the issue of investigating the (long-term) memorability of model $\Theta$ can be converted into a new problem of statistically testing the mismatch between the transformed and observed processes. We design the corresponding numerical experiments as follows.
\begin{itemize}
	\item \textbf{Step 1.} Define the averaged memory statistic $\bar{d}$ described in Subsection~\ref{subsec:EMH}.
	\item \textbf{Step 2.} For the task of estimating the long-term memorability of the conducted model $\Theta$, we feed up the model $\Theta$ with the observations $\{ \mathbf{X}_t \}_{t=0}^{T}$ of differencing indicator $\bar{d}^* \geq 0$, and then, calculate the residual statistic $\bar{d}-\bar{d}^*$, which indicates the gap between the memorability of model $\Theta$ and long-term dependent characteristic of the conducted data. 
	\item \textbf{Step 3.} The testable criteria becomes to test the null hypothesis $\mathcal{H}_0: \bar{d}-\bar{d}^* = 0$ the one-side alternative of long memory $\mathcal{H}_1: \bar{d}-\bar{d}^* < 0$, which corresponds to the model' failure to represent the full strength of long-term memory observed from the sequential data.
\end{itemize}

Note that we here present the residual statistic $\bar{d}-\bar{d}^*$ to investigate the the memorability of model $\Theta$. If $\{ \mathbf{X}_t \}_{t=0}^{\infty}$ belongs to a long-range dependent process (i.e., $\bar{d}^* > 0$), a non-zero residual statistic reveals a mismatch between representation structure learned by the ARISE-$\Theta$ process and the long-range dependency of $\{ \mathbf{X}_t \}_{t=0}^{T}$. On the other hand, if $\mathbf{X}_t$ belongs to the Gaussian white noise, the non-zero residual statistic corresponds the long-term temporal representation of the ARISE-$\Theta$ process such that zero-memory inputs are transformed to a long-range dependent sequence.

The experiment inherits the data sets used in Subsection~\ref{subsec:EMH}. The conducted models contain vanilla RNN~\citep{pearlmutter1995}, MGU~\citep{zhou2016}, GRU~\citep{cho2014}, LSTM~\citep{hochreiter1997}), and FTNet~\citep{zhang2021}. These models use one hidden layer with 150 neurons and are trained by the Adam algorithm~\citep{kingma2014:adam} within 200 epochs. We also run each experiment 5 times for counting the mean and variance of the residual statistic. The experimental results are listed in Table~\ref{tab:applications}, from which each model fails to pursue the (long-/short-range) statistical dependency characteristics of the conducted data sets, including \texttt{CSI 300 Index}, \texttt{Winton Stock Exchange}, \texttt{SSEC}, and \texttt{Penn TreeBank}. These results shed some insights on a standpoint that the well-known machine learning models do not have long-term temporal representation in a statistical sense.

\subsection{Latent State-Space Model for Inference and Forecasting} \label{subsec:ldss}
\begin{figure}[t]
	\centering
	\includegraphics[width=0.75\textwidth]{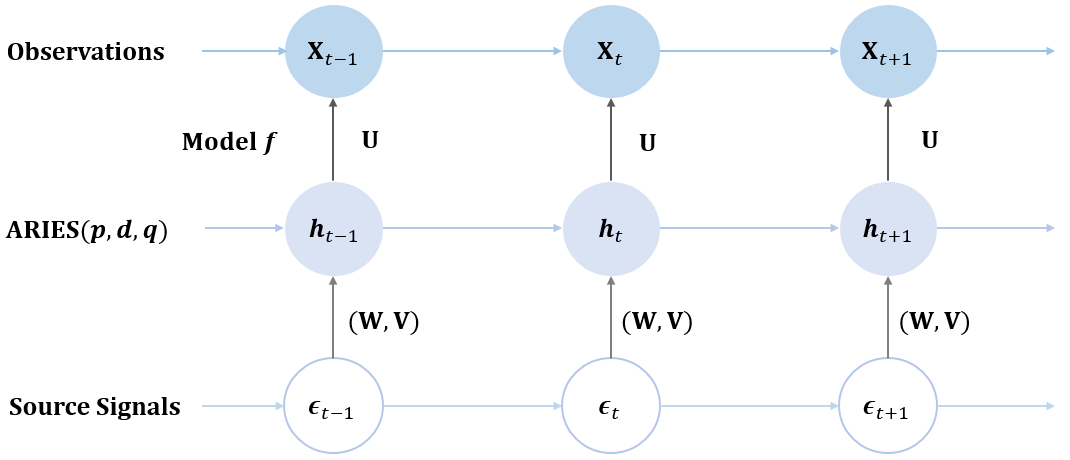}
	\caption{Topology illustration of the LDSS model.}
	\label{fig:LDSS}
\end{figure}

The past decades have witnessed the glories and success of the \textit{latent state-space} models in abundant fields, such as uncertainty estimation~\citep{gupta2006:ue,gylys2020:ue}, casual inference~\citep{barnett2015:causal,huang2019:causal,pearl2009:causal}, control systems~\citep{kalman1960:control,valpola2002:control}, etc. Traditional latent state-space models perform effectively on modeling systems where the dynamics are approximately linear and relatively simple, whereas challenging for capturing the temporal structure of time series with long-term dependency in terms of both inference and forecasting. Here, we present the ARISE-based \textit{Long-Dependent State-Space} (LDSS) model for time series probability analysis, which utilizes the ASE estimation as priors for time series with long-term dependency and employs the ARISE model to mimic the latent state trajectories. Figure~\ref{fig:LDSS} illustrates its topology. Given the observations $\{\mathbf{X}_t\}_{t=1}^T$ in the training range $[T]$, the procedure for inference and forecasting is listed as follows
\begin{itemize}
	\item \textbf{Step 1.} Calculating the estimation value $\widehat{\boldsymbol{d}}_{\mathrm{ASE}}$ for the observations $\{\mathbf{X}_t\}_{t=1}^T$.
	\item \textbf{Step 2.} Generating the latent state sequences $\{ \boldsymbol{h}_t \}_{t=1}^{\infty}$ via the ARISE-ARMA$(p,\widehat{\boldsymbol{d}}_{\mathrm{ASE}},q)$ model
	\begin{equation} \label{eq:vector_fd}
		\left( \mathbf{W}_0 + \sum_{k=1}^{p} \mathbf{W}_k \mathfrak{B}^k \right)	\begin{pmatrix}
			(1-\mathfrak{B})^{d_1} & & 0 \\
			& \ddots & \\
			0 & & (1-\mathfrak{B})^{d_l}
		\end{pmatrix} \boldsymbol{h}_t = \left( \mathbf{V}_0 + \sum_{k=1}^{q} \mathbf{V}_k \mathfrak{B}^k \right) \boldsymbol{\epsilon}_t ,
	\end{equation}
	where $(p,q)$ is a pair of hyper-parameters and the source signals $\{ \boldsymbol{\epsilon}_t \}_{t=1}^{\infty} \sim \mathcal{N}(\boldsymbol{\mu},\Sigma_{\epsilon})$.
	\item \textbf{Step 3.} Suppose $\mathbf{X}_t \sim \mathcal{N} \left( \mathbf{U}\boldsymbol{h}_t, \Sigma_h \right)$ and let $\theta = \left( \boldsymbol{\mu}, \Sigma_{\epsilon}, \mathbf{W}_{0:p}, \mathbf{V}_{0:q}, \mathbf{U}, \Sigma_h \right)$ fully specify the parameters. Generally, we maximize the marginal likelihood $\theta^* = \arg\max_{\theta} \mathcal{P}_{ldss} \left( \mathbf{X}_{1:T} \mid \theta \right)$, where for $r = \max\{p,q\}$,
	\[
	\begin{aligned}
		\mathcal{P}_{ldss} \left( \mathbf{X}_{1:T} \mid \theta \right) &= p \left( \mathbf{X}_{1:r} \mid \theta \right) \!\!\! \prod_{t=r+1}^{T} \!\! p \left( \mathbf{X}_t \mid \mathbf{X}_{1:t-1}, \theta \right) \\
		&= \int p(\boldsymbol{h}_{1:r}) \left[ \prod_{t=r+1}^{T} p \left( \mathbf{X}_t \mid \boldsymbol{h}_t \right) p \left( \boldsymbol{h}_t \mid \boldsymbol{h}_{t-1} \right) \right] \dif \boldsymbol{h}_{1:T}
	\end{aligned}
	\]
	denotes the marginal probability of the observations $\{\mathbf{X}_t\}_{t=1}^T$ given the latent state sequence $\{ \boldsymbol{h}_t \}_{t=1}^T$ and parameters $\theta$. Practically, we recommend $r \leq 1$ as the default and employ the TVAR inference algorithm~\citep{west1999:TVAR} for accelerating optimization.
	\item \textbf{Step 4.} Provided $\theta^*$, we proceed to predict probabilistic forecasts according to 
	\[
	p \left( \mathbf{X}_{T+1} \mid \mathbf{X}_{1:T},  \boldsymbol{\epsilon}_{1:T+1}, \boldsymbol{h}_{1:T+1}; \theta^* \right) ,
	\]
	from which we can analytically compute the joint distribution over the prediction range for each time series as this joint distribution is a multivariate Gaussian. Here, we use $K$ Monte Carlo samples to indicate the forecast distribution $\widehat{\mathbf{X}}_{T+1}^{(k)} \sim p \left( \mathbf{X}_{T+1} \mid \mathbf{X}_{1:T}, \boldsymbol{\epsilon}_{1:T+1}, \boldsymbol{h}_{1:T+1}; \theta^* \right)$ for $k \in[K]$.
\end{itemize}

Notice the fractional differencing calculation of Eq.~\eqref{eq:B_extension} in Step 2. Let $\{ h_t \}_{t=1}^T$ denote the concerned time series. For the example of lag $k=4$, we have
\[
\left\{\begin{aligned}
	(1-\mathfrak{B})^0 h_t &= 1 * h_t + 0 * h_{t-1} + 0 * h_{t-2} + 0 * h_{t-3} + 0 * h_{t-4} ,\\
	(1-\mathfrak{B})^{0.4} h_t &= 1 * h_t - 0.4 * h_{t-1} - 0.12 * h_{t-2} - 0.064 * h_{t-3} - 0.0416 * h_{t-4} ,\\
	(1-\mathfrak{B})^1 h_t &= 1 * h_t - 1 * h_{t-1} + 0 * h_{t-2} + 0 * h_{t-3} + 0 * h_{t-4} .\\
\end{aligned}\right.
\]
It's observed that the coefficient absolutely converges to zero as the lag $k$ goes to infinity. Thus, it is completely reasonable to use the finite truncation to approximate the fractional differencing value. Here, we recommend the lag $k=4$ as a default scalar. Therefore, the fractional differencing term in Eq.~\eqref{eq:vector_fd} can be rewritten as
\[
\begin{pmatrix}
	(1-\mathfrak{B})^{d_1} & & 0 \\
	& \ddots & \\
	0 & & (1-\mathfrak{B})^{d_l}
\end{pmatrix} \boldsymbol{h}_t = \begin{pmatrix}
	(1-\mathfrak{B})^{d_1} \boldsymbol{h}_{1t} \\
	\vdots\\
	(1-\mathfrak{B})^{d_l} \boldsymbol{h}_{lt} 
\end{pmatrix} \approx \begin{pmatrix}
	1 & b_{11} & b_{12} & b_{13} & b_{14}\\
	\vdots &  & \ddots &  & \vdots \\
	1 & b_{l1} & b_{l2} & b_{l3} & b_{l4}\\
\end{pmatrix} \begin{pmatrix}
	\boldsymbol{h}_{1t} \\
	\vdots\\
	\boldsymbol{h}_{lt} 
\end{pmatrix} ,
\]
where the row vector $(1, b_{i1}, b_{i2}, b_{i3}, b_{i4} )$ denotes the fractional differencing coefficients for $i \in [l]$.

\begin{table}
	\centering
	\caption{Forecasting performance of the LDSS model and its comparative models.}
	\label{tab:forecasting}
	\resizebox{\textwidth}{!}{
		\begin{tabular}{@{}ccc|ccc|ccc@{}}
			\toprule[2pt]
			Data Sets & Models & MSE & Data Sets & Models & MSE ($10^{-2}$) & Data Sets & Models & MSE ($10^5$) \\ 
			\midrule[1.5pt]
			\multirow{7}{*}{\texttt{Electricity}}
			& ARIMA(8,1,4) & 1.3480
			& \multirow{7}{*}{\makecell[c]{\texttt{Exchange}\\ \texttt{Rate}}} 
			& ARIMA(8,1,4) & $8.2790$ 
			& \multirow{7}{*}{\makecell[c]{\texttt{Yancheng}\\ \texttt{Automobile}\\ \texttt{Registration}}} 
			& ARIMA(6,1,3)  & $84.5129$ \\
			& KNNs(1,1) & 1.0824
			&
			& KNNs(1,1) & $7.9351$
			&
			& KNNs(1,1)    & $31.2573$ \\
			& TVAR ($k=5$) & 0.6453
			&
			& TVAR($k=7$)  & $7.9235$
			&
			& NARXnet & $20.2631$ \\
			& GRU  & 0.8731
			&
			& GRU & $6.2748$
			&
			& GRU & $13.0421$ \\
			& DeepAR & 0.5233
			&  
			& DeepAR & $5.2074$
			&
			& DeepAR & $10.7250$ \\
			& LSTNet & 0.6427
			& 
			& LSTNet & $5.1732$
			& 
			& LSTNet & $8.4176$ \\
			& LDSS($k=4$) & $\mathbf{0.4273}$ 
			&
			& LDSS($k=4$) & $\mathbf{5.0262}$
			&
			& LDSS(k=4) & $\mathbf{8.2970}$ \\
			\bottomrule[2pt]
	\end{tabular} }
\end{table}
The first experiment is to demonstrate the forecasting performance of our proposed LDSS model on three real-world data sets. The data sets we consider consist of (1) \texttt{Electricity}~\footnote{\url{https://archive.ics.uci.edu/ml/datasets/ElectricityLoadDiagrams20112014}}, hourly electricity consumption for 370 customers from 2012 to 2014, (2) \texttt{Exchange-Rate}, the collection of the daily exchange rates of eight foreign countries including Australia, British, Canada, Switzerland, China, Japan, New Zealand and Singapore ranging from 1990 to 2016~\citep{lai2018:Exchange-Rate}, and (3) \texttt{Yancheng Automobile Registration}~\footnote{\url{https://tianchi.aliyun.com/competition/entrance/231641/information}}, the registration records of 5 car brands in nearly 1000 dates. All data sets have been split into training set (60\%), validation set (20\%), and test set (20\%) in chronological order. The contenders contains ARIMA, KNNs~\citep{yang2011:knn}, TVAR, and several deep-learning models, like DeepAR~\citep{salinas2020:deepar}, LSTNet~\cite{lai2018:Exchange-Rate}, NARXnet~\citep{guzman2017:narx}, etc. Table~\ref{tab:forecasting} lists the forecasting results and shows that the LDSS model achieve the best performance, marked in bold, against its comparative models on three real-world data sets.

\begin{figure}[t]
	\centering
	\includegraphics[width=0.9\textwidth]{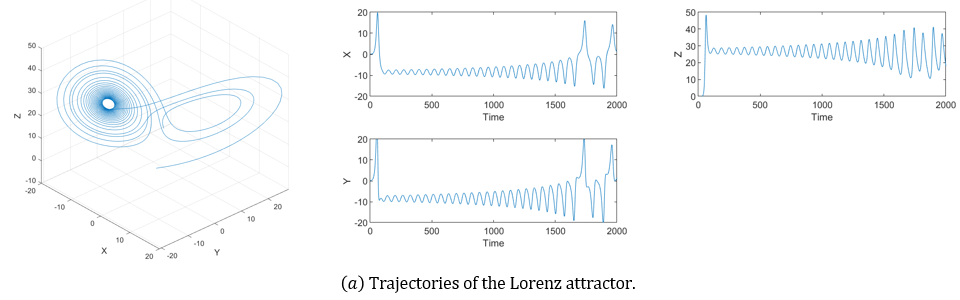}
	\includegraphics[width=0.9\textwidth]{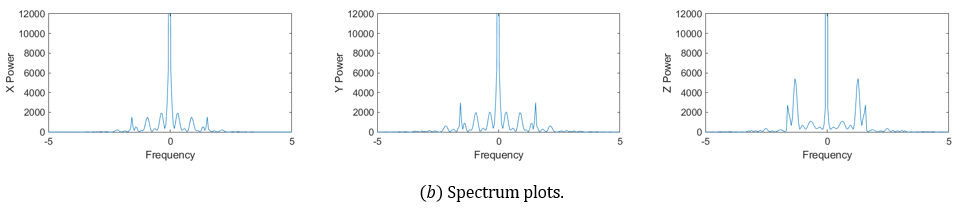}
	\includegraphics[width=0.9\textwidth]{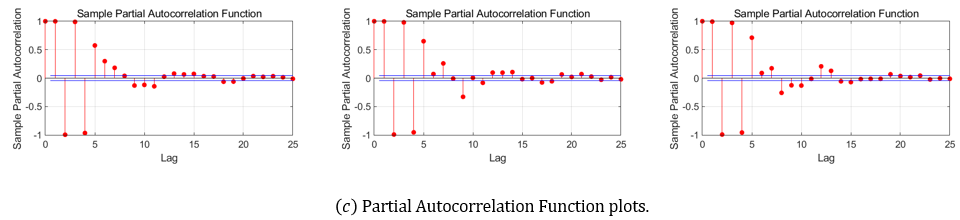}
	\caption{The simulated trajectories, spectrum, and partial auto-correlation plots of $x,y,z$ of a Lorenz attractor.}
	\label{fig:simulation}
\end{figure}

The second experiment is to demonstrate the probabilistic inference of LDSS for mimicing complex physical systems. We here simulate 2000 points from the following Lorenz attractor with a timestamp of 0.01,
\[
\frac{\dif x}{\dif t} = a(y-x) \quad\quad
\frac{\dif y}{\dif t} = x(b-z)-y \quad\quad
\frac{\dif z}{\dif t} = x y-cz ,
\]
\begin{wraptable}{r}{0.5\textwidth}
	\centering
	\caption{CP of the DLM, TVAR, and LDSS models on simulated Lorenz attractor.}
	\label{tab:coverage}
	\resizebox{0.5\textwidth}{!}{
		\begin{tabular}{@{}cccc@{}}
			\toprule[2pt]
			Models & CP ($68.27\%$) & CP ($95.45\%$)  & CP ($99.73\%$) \\
			\midrule[1.5pt]
			DLM   & $79.88\%$ & $96.68\%$ & $99.78\%$ \\ 
			TVAR  & $82.41\%$ & $97.17\%$ & $\mathbf{99.95\%}$ \\ 
			LDSS  & $\mathbf{86.54\%}$ & $\mathbf{97.85\%}$ & $\mathbf{99.95\%}$ \\ 
			\bottomrule[2pt]
	\end{tabular} }
\end{wraptable}
where $a,b,c$ are real-valued scalars and we adopt $(a,b,c) = (10,28,8/3)$ as the default. Figure~\ref{fig:simulation} displays the trajectories of variables $x,y,z$, which have more irregular marginals (heavier-tailed, persistent, etc.) than the Gaussian ones. We fit this system by using the LDSS model, and then, output the mean trajectories and posterior forecast marginals of the learned LDSS model. We expect the simulated points to fall within the credible intervals for the posterior forecast marginals of the learned LDSS model. For precision, we calculate the coverage percentage of simulated points that fall within specific percentiles (e.g., $95.45\%$) of the posterior forecast marginals, and employ this \textit{coverage percentage} (CP) as a new-born evaluation indicator for demonstrating the fitting performance of the LDSS and its contenders, that is, TVAR and DLM~\citep{gruber2016:dlm}. Table~\ref{tab:coverage} lists the coverage percentages of the DLM, TVAR, and LDSS models against the theoretical percentiles $68.27\%$, $95.45\%$, and $99.73\%$, which confirms that the proposed LDSS model outperforms the competition.

In summary, we develop the latent state-space model using the ARISE process for both inference and forecasting of time series. The experiments evaluate its inference utility for retrieving the underlying structure of chaos systems and its competitive forecasting performance to several state-of-the-art models on time series with long-term dependence.

\section{Conclusions, Discussions, and Prospects} \label{sec:conclusions}
In this paper, we proposed the ARISE process for investigating the efficient market hypothesis. The ARISE process is a semi-parametric approach that consists of a parametric integrated process with an infinite-sum function of some known processes and the non-parametric ASE based on apposite wavelet-threshold methods, thus with the power and potential of modeling the price data with long-term memory, non-stationarity, and aperiodic spectrum. We theoretically establish the well-posed properties, such as the mean-square convergence, consistency, and asymptotic normality, of the ARISE process without assuming periodogram and Gaussianity. We use the ARISE process in reality to determine the efficiency of real-world markets, study the long-term memorability of multiple machine-learning models, and construct a latent state-space model for inference and forecasting. The numerical experiments confirm the superiority of our proposed approaches.

Long-term memory is perhaps the most important characteristic of efficient markets. The current solution is to take the considerably simple form, i.e., connect the values at all accessible timestamps by a parametric integrated process, for mimicking the long-term dependency between observations. Such a modeling approach has two favorite properties. First, the relationship between the value of the memory parameter and the persistence of a shock is easily understood from the inverse expansion, described in Eq.~\eqref{eq:B_extension}, albeit this is only formal for $d_i > 0$. Second, using the coefficients in Eq.~\eqref{eq:B_extension}, one can focus microscopical attention on the observation at any timestamp. On the contrary, such a model unavoidably necessitates a huge calculation, as introduced in Section~\ref{sec:ARISE}. Besides, this model may be not good at typical machine learning prediction tasks, which require balancing the memory of commonality and the forgetting of particularity, whereas the ARISE process pays too much attention to micro-features of the observations. 

In light of the preceding merits and defects, we feel the ARISE process is more suitable for jobs that require both macro-/micro-cosmic modelings as well as comprehensibility. Such tasks include the efficient market hypothesis that this work focuses on and the uncertainty estimation, causal inference, and control strategy in other fields, such as hydrology, language, meteorology, etc. In the future, it is interesting to explore the more flexible implementation and a broader range of applications for machine learning and other fields.

\section*{Acknowledgments}
This research was supported by the National Science Foundation of China (61921006) and the Program B for Outstanding Ph.D Candidate of Nanjing University (202101B051). The authors would like to thank the anonumous reviewers for constructive suggestions, as well as Zhen-Yu Zhang for helpful discussions.


\newpage
\appendix
\begin{center}
	\Large\textbf{Supplementary Materials of the ARISE Process (Appendix)}
\end{center}

\section{Technical Lemmas} \label{app:TL}	
Here, we list some useful techniques.
\begin{lemma}[Lemma 1 in \citep{robinson1995gaussian}] \label{tec:robinson}
	For $\epsilon \in (0,1)$ and $C \in (\epsilon, \infty)$, we have
	\[
	\sup_{C \geq \gamma \geq \varepsilon} \left| \frac{\gamma}{m} \sum_{j=1}^{m} \left(\frac{j}{m}\right)^{\gamma-1}-1 \right| = \mathcal{O} \left( \frac{1}{m^{\varepsilon}} \right) \quad\text{as}\quad m \to \infty.
	\]
\end{lemma} 

\begin{lemma}[Lemma 2 in \citep{robinson1995gaussian}] \label{tec:robinson:2}
	For $m \geq 2$, we have
	\[
	\left|\frac{1}{m} \sum_{1}^{m} \log j-\log m+1\right| \leq \frac{2+\log (m-1)}{m} .
	\]
\end{lemma}

\begin{lemma}[Lemma 5.4 in \citep{shimotsu2005exact}] \label{tec:shimotsu2005}
	For $\kappa \in (0,1)$ and $C \in (1, +\infty)$, the followings hold
	\begin{itemize}
		\item[i)] $\sup_{\gamma \in [-C,C]} \left| m^{-1} \sum_{j=1}^{m} (j/m)^{\gamma} - \int_{\kappa}^1 x^{\kappa} \dif x \right| = \mathcal{O} \left( m^{-1} \right)$ as $m \to \infty$;
		\item[ii)] $\liminf_{m \to \infty} \inf_{\gamma \in [-C,C]} \left( m^{-1} \sum_{j=\lfloor m\kappa \rfloor}^{m} (j/m)^{\gamma} \right) > 0 $ \\ and 
		$\sup_{\gamma \in [-C,C]} \left| m^{-1} \sum_{j=\lfloor m\kappa \rfloor}^{m} (j/m)^{\gamma} \right| = \mathcal{O} \left( 1 \right)$ as $m \to \infty$.
	\end{itemize}
\end{lemma}

\begin{lemma} \label{tec:our1}
	As $T \to \infty$, we have
	\[
	\frac{\sqrt{m} \log m \log T}{T} = \frac{\sqrt{m}}{\sqrt{T}} \frac{\log (m)}{\sqrt[4]{T}} \frac{\log T}{\sqrt[4]{T}} \to 0 .
	\]
\end{lemma}

\begin{lemma} \label{tec:an}
	Let $\{\mathbf{X}_t\}_{t=0}^{\infty}$ be a $l$-dimensional process specified by Eq.~\eqref{eq:starting}, which meets Assumption~\ref{chara:1}, $f_X$ is the corresponding spectral density matrix, $\mathbf{I}_T$ and $\mathbf{J}_T$ are empirical spectral density estimators, described by Eqs.~\eqref{eq:I} and~\eqref{eq:J}, respectively. For $p,q \in[l]$, $\phi_{p,q}$ and $\phi_{p,q}'$ denote the $(p,q)$-element of matrices $\mathbf{I}_T$ and $\mathbf{J}_T$, respectively. Let Assumptions~\ref{ass:3},~\ref{ass:B1}-\ref{ass:B4} holds. We have
	\begin{itemize}
		\item[i)] $ \max_{p,q\in[l]} \left\{ \sum_{j=1}^{m} \left( \phi_{p,q} - \phi'_{p,q} \right) \right\} = \op \left( {m}/{T^{1+|d^0_p+d^0_s|}} \right) $;
		\item[ii)] $\max_{p,q\in[l]} \left\{ \sum_{j=1}^{m} \e^{\ii (\lambda_j - \pi) (d_p^0-d_q^0)/2} \lambda_j^{d_p^0+d_q^0} \left( \phi_{p,q} - \phi'_{p,q} \right)  \right\} = \op\left( \sqrt{m}/\log m \right)$;
		\item[iii)] $\max_{p,q\in[l]} \left\{ \sum_{j=1}^{m} \e^{\ii (\lambda_j - \pi) (d_p^0-d_q^0)/2} \lambda_j^{d_p^0+d_q^0} \phi_{p,q} - \phi'_{p,q} - G^0_{p,q}  \right\} = \op\left( {m^{\alpha}}/{T^{\alpha}} + \sqrt{m}\log m \right)$.
	\end{itemize}
\end{lemma}

\begin{lemma} \label{tec:our2}
	Let $a_j^{(k)}(\boldsymbol{d}) = (\log \lambda_j)^k \e^{\ii (\lambda_j-\pi)(d_p-d_q)/2} \lambda_j^{d_p+d_q}$ and $\Omega_{\delta,T} = \left\{ \boldsymbol{d} \mid (\log T)^4 \| \boldsymbol{d} - \boldsymbol{d}^0 \|_{\infty} < \delta \right\}$ for $\delta>0$, $k=0,1,2$, and $p,q \in [l]$. Then for $\boldsymbol{d} \in \Omega_{\delta,T}$, there exists a constant $C > 1$ such that
	\[
	a_j^{(k)}(\boldsymbol{d}) - a_{j+1}^{(k)}(\boldsymbol{d}) = \mathcal{O} \left( (\log T)^k j^{-1} \right) ,\quad
	a_m^{(k)}(\boldsymbol{d}) = \mathcal{O} \left( (\log T)^k \right)  ,\quad
	\left| a_j^{(0)}(\boldsymbol{d}) \lambda_j^{-d_p-d_q} - 1 \right| \leq C(|d_p|+|d_q|) . 
	\]
	Furthermore, we have
	\[
	\left| \frac{\lambda_j^{d_p+d_q} - 1}{d_p+d_q} \right| \leq | \log \lambda_j | T^{|d_p|+|d_q|} \leq C\log T
	\quad\text{and}\quad
	\frac{1}{m} \sum_{j=1}^{m} \left( \log\lambda_j \right)^2 - \left( \frac{1}{m} \sum_{j=1}^{m} \log \lambda_j \right)^2 = \Theta(1) .
	\]
\end{lemma}

\section{Full Proofs for Theorem~\ref{thm:convergence}} \label{app:convergence}
Before our proof, we should introduce some definitions and lemmas.
\begin{definition}[From \citep{fuhr2010:besov,machihara2003:besov}]
	The Besov space\\ $\mathcal{B}_{p,q}^n(\mathbb{R};R)$ is a quasi-normed space with $n,p,q \leq1 $ and $R>0$
	\[
	\mathcal{B}_{p,q}^n(\mathbb{R};R) = \left\{ f ~\bigg|~ \|f\|_{\mathcal{B}_{p,q}^n(\mathbb{R})} \overset{\underset{\mathrm{def}}{}}{=} \left( \|f\|_{\mathcal{W}^{n, p}(\mathbb{R})}^q + \int_{0}^{\infty} \left| \frac{\lambda_p^2\left(f^{(n)}, t\right)}{t^{\alpha}} \right|^q \frac{\dif t}{t}\right)^{\frac{1}{q}} \leq R \right\} .
	\]
\end{definition}
\noindent\textbf{Inconsistency of $\mathbf{I}'_T(\lambda_j)$.} Recall the tapered periodogram estimation for $f_X(\lambda)$ in Eq.~\eqref{eq:J'} and observe that the tapper function $\hbar_i$ is of bounded variation and satisfies $\int_{0}^{1} \hbar_i(x) \dif x >0$. Thus, the tapered periodogram $\mathbf{I}'_T(\lambda_j)$ in Eq.~\eqref{eq:J'} can be rewritten as
\[
\mathbf{I}'_T(\lambda_j) = \frac{1}{2\pi \sum_{s=1}^{T} \hbar^2(s/T)} \left| \sum_{s=1}^{T} \sum_{t=1}^{T} \hbar\left( s/T \right) \hbar\left( t/T \right) \langle \mathbf{X}_t, \mathbf{X}_s \rangle \e^{\ii (t-s) \lambda_j} \right|^2 .
\] 
It's well known that $\mathbf{I}'_T$ is an inconsistent estimator for $f_X$ and needs to be smoothed to achieve consistency. 

\noindent\textbf{Thresholds Construction for $\boldsymbol{\rho_{i,\kappa}}$.} Inspired by Donoho~\citep{donoho1994}, the desired thresholds $\rho_{i,\kappa}$ should be an ``apposite" approximation to the local weighted $l_1$ norm of the periodogram $\mathbf{I}_T(\lambda)$ corresponding to $f_X(\lambda)$. Thus, the desired thresholds should be in the following form
\begin{equation} \label{eq:rho}
	\rho_{i,\kappa} = \widehat{C} \cdot \left( \int_{-\pi}^{\pi} \varphi_{i,\kappa}(\lambda) f_X(\lambda) \dif \lambda \right) \cdot \sqrt{2 \log(|\mathfrak{J}_T|_{\#}) } ~,\quad\text{for ~some}~~ \widehat{C}, 
\end{equation}
where the basis function $\varphi_{i,\kappa}(\lambda) = 2^{i/2} \varphi(2^i \lambda - \kappa)$ is non-negative on $L_2[-\pi,\pi]$ normalized to square-integrate to one, which satisfies that $\int_{-\pi}^{\pi} \varphi_{0,\kappa}(\lambda) \dif \lambda =1$ and $\int_{-\pi}^{\pi} \varphi_{i,\kappa}(\lambda) \lambda^s \dif \lambda =0$ for $i \neq 0$ and $s\in \mathbb{N}^+$. Since the approximation scalar $\widehat{C} \int_{-\pi}^{\pi} \varphi_{i,\kappa}(\lambda) f_X(\lambda) \dif \lambda$ indicates the local weighted $l_1$ of $f_X(\lambda)$ at scale $i$ and location $\kappa$, we here set $\widehat{C} = \Theta(T^{-1/2})$, and thus, obtain 
\[
\sup_{(i,\kappa) \in \mathfrak{J}_T} \rho_{i,\kappa} = \mathcal{O} ( \sqrt{ {\log T}/{T} } ) .
\]
However, $\rho_{i,\kappa}$ in Eq.~\eqref{eq:rho} with $\widehat{C} = \Theta(T^{-1/2})$ is still impractical as it involves the underlying $f_X(\lambda_j)$. To tackle this challenge, in Algorithm~\ref{alg:ASE}, we practically calculate the threshold $\rho_{i,\kappa}$, descried in Eq.~\eqref{eq:rho}, by this following approximated one, described in Eq.~\eqref{eq:rho_text},
\[
\widehat{\rho}_{i,\kappa} = \Theta\left( T^{-1/2} \right) \left( \int_{-\pi}^{\pi} \varphi_{i,\kappa}(\lambda) \mathbf{I}_T(\lambda) \dif \lambda \right) ~ \sqrt{2 \log(|\mathfrak{J}_T|_{\#}) } ~.
\]
It suffices to show that $\widehat{\rho}_{i,\kappa}$ is a ``well-defined approximator" for  $\rho_{i,\kappa}$. 
From \citep[Proposition 3.1]{neumann1996}, we have
\begin{equation} \label{eq:thm_use1}
	\widehat{C} \int_{-\pi}^{\pi} \varphi_{i,\kappa}(\lambda) f_T(\lambda) \dif \lambda = \Theta \left( T^{-1/2} \right) \int_{-\pi}^{\pi} \varphi_{i,\kappa}(\lambda) f_T(\lambda) \dif \lambda = \mathcal{O} \left( 2^{i/2} T^{-1} \right)
\end{equation}
and
\[
\mathbb{E} \left[ \int_{-\pi}^{\pi} \varphi_{i,\kappa}(\lambda) \mathbf{I}_T(\lambda) \dif \lambda \right] - \int_{-\pi}^{\pi} \varphi_{i,\kappa}(\lambda) f_T(\lambda) \dif \lambda = \mathcal{O} \left( 2^i T^{-1} \log T  \right) ,
\]
which means that $\widehat{C}^2 \int_{-\pi}^{\pi} f_X^2(\lambda) \varphi_{i,\kappa}(\lambda) \left( \varphi_{i,\kappa}(\lambda) + \varphi_{i,\kappa}(-\lambda) \right) \dif \lambda$ is of order $\mathcal{O}(2^i T^{-1})$ and bias is of order $\mathcal{O}( 2^i T^{-1} \log T )$. Let
\[
\sigma_{i,\kappa} = \mathrm{VARIANCE}\left( \widehat{C} \left( \int_{-\pi}^{\pi} \varphi_{i,\kappa}(\lambda) \mathbf{I}_X(\lambda) \dif \lambda \right) \right).
\]
From \citep[Lemma 6]{dahlhaus1983}, it is observed that
\begin{equation} \label{eq:thm_use2}
	\begin{aligned}
		\sigma^2_{i,\kappa} &= \widehat{C}^2 \left[ \widehat{C}^2 \int_{-\pi}^{\pi} f_X^2(\lambda) \varphi_{i,\kappa}(\lambda) \left( \varphi_{i,\kappa}(\lambda) + \varphi_{i,\kappa}(-\lambda) \right) \dif \lambda + o(2^i T^{-1}) + \mathcal{O}(T^{-1}) \right] \\
		&= \widehat{C}^2 \left[ \mathcal{O}(2^i T^{-1}) + o(2^i T^{-1}) + \mathcal{O}(T^{-1}) \right] .
	\end{aligned}
\end{equation}
Thus, we have  
\[
\resizebox{1\hsize}{!}{$
	\begin{aligned}
		&\sum_{(i,\kappa) \in \mathfrak{J}_T} \mathbb{P} \left( \widehat{\rho}_{i,\kappa} < \rho_{i,\kappa} \right)  \\
		&\quad= \sum_{(i,\kappa) \in \mathfrak{J}_T} \mathbb{P} \left( \frac{ C \int_{-\pi}^{\pi} \varphi_{i,\kappa}(\lambda) \left( f_X(\lambda) - \mathbf{I}_T(\lambda) \right) \dif \lambda }{\sigma_{i,\kappa}} > \frac{ C (1 - \mu(T)) \int_{-\pi}^{\pi} \varphi_{i,\kappa}(\lambda) f_X(\lambda) \dif \lambda }{\sigma_{i,\kappa}} \right)  \\
		&\quad\leq \sum_{(i,\kappa) \in \mathfrak{J}_T} \mathbb{P} \left( \frac{ C \int_{-\pi}^{\pi} \varphi_{i,\kappa}(\lambda) \left( f_X(\lambda) - \mathbf{I}_T(\lambda) \right) \dif \lambda }{\sigma_{i,\kappa}} > \min\left\{ \frac{ C (1 - \mu(T)) \int_{-\pi}^{\pi} \varphi_{i,\kappa}(\lambda) f_X(\lambda) \dif \lambda }{\sigma_{i,\kappa}} , \nu(T) \right\} \right)  \\
		&\quad\leq C_1 \left[ \sum_{(i,\kappa) \in \mathfrak{J}_T} \mathbb{P} \left( Z > \frac{ C (1 - \mu(T)) \int_{-\pi}^{\pi} \varphi_{i,\kappa}(\lambda) f_X(\lambda) \dif \lambda }{\sigma_{i,\kappa}} \right) + \sum_{(i,\kappa) \in \mathfrak{J}_T} \mathbb{P} \left( Z > \nu(T) \right) \right]  \\
		&\quad\leq C_1 \left[ \sum_{(i,\kappa) \in \mathfrak{J}_T} \mathbb{P} \left( Z > C_2 (1 - \mu(T)) 2^{(i'-i)/2} \right) + \sum_{(i,\kappa) \in \mathfrak{J}_T} \mathbb{P} \left( Z > \nu(T) \right) \right] 
		\quad\left( \text{Invoking Eqs.~\eqref{eq:thm_use1} and~\eqref{eq:thm_use2} } \right) \\
		&\quad\leq C_1 \left[ \sum_{(i,\kappa) \in \mathfrak{J}_T} e^{-C_2^2 2^{i'-i}(1-\mu(T))^2/2} + \sum_{(i,\kappa) \in \mathfrak{J}_T} \e^{- \nu(T)^2/2} \right] 
		\quad\left( \text{From Azuma's inequality} \right) ,\\
	\end{aligned} $}
\]
where $C_1, C_2, C_3 >0$ and
\[
\begin{aligned}
	& C=\widehat{C}\sqrt{2 \log(|\mathfrak{J}_T|_{\#}) } ,\\
	& 1 - \mu(T) \geq T^{-\delta / 2} \sqrt{C_3(1-\delta) \log T - T} ,\\
	& \nu(T) = T^{\delta/2(3+4\gamma)}  (\log T)^{-1-1/(3+4\gamma)} .
\end{aligned}
\]
On the other hand, we have
\[
\resizebox{1\hsize}{!}{$
	\begin{aligned}
		&\sum_{(i,\kappa) \in \mathfrak{J}_T} \mathbb{P} \left( \widehat{\rho}_{i,\kappa} > \rho_{i,\kappa} \right)  \\
		&\quad= \sum_{(i,\kappa) \in \mathfrak{J}_T} \mathbb{P} \left( \frac{ C \int_{-\pi}^{\pi} \varphi_{i,\kappa}(\lambda) \left( f_X(\lambda) - \mathbf{I}_T(\lambda) \right) \dif \lambda }{\sigma_{i,\kappa}} > \frac{ C_4 - C \int_{-\pi}^{\pi} \varphi_{i,\kappa}(\lambda) f_X(\lambda) \dif \lambda }{\sigma_{i,\kappa}} \right)  \\
		&\quad\leq \sum_{(i,\kappa) \in \mathfrak{J}_T} \mathbb{P} \left( \frac{ C \int_{-\pi}^{\pi} \varphi_{i,\kappa}(\lambda) \left( f_X(\lambda) - \mathbf{I}_T(\lambda) \right) \dif \lambda }{\sigma_{i,\kappa}} > C_5T^{1/2}2^{(i'-i)/2} \right)
		\quad\left( \text{Invoking Eqs.~\eqref{eq:thm_use1} and~\eqref{eq:thm_use2} } \right) \\
		&\quad\leq C_1 \sum_{(i,\kappa) \in \mathfrak{J}_T} \mathbb{P} \left( Z > \nu(T) \right) \\
		&\quad\leq C_1 \sum_{(i,\kappa) \in \mathfrak{J}_T} \e^{- \nu(T)^2/2}
		\quad\left( \text{From Azuma's inequality} \right) ,\\
	\end{aligned} $}
\]
where $C_4, C_5 >0$. Thus, $\widehat{\rho}_{i,\kappa}$ is a consistent estimator for  $\rho_{i,\kappa}$, i.e.,
\[
\sum_{(i,\kappa) \in \mathfrak{J}_T} \mathbb{P} \left( \widehat{\rho}_{i,\kappa} < \rho_{i,\kappa} \right) \to 0
\quad\text{and}\quad
\sum_{(i,\kappa) \in \mathfrak{J}_T} \mathbb{P} \left( \widehat{\rho}_{i,\kappa} > \rho_{i,\kappa} \right) \to 0
\quad\text{as}\quad T \to \infty  .
\]

\noindent\textbf{Near-optimal rate of mean-square convergence.} Here, we consider the concerned spectral density $f_X(\lambda)$ lies in a Besov ball $\mathcal{B}_{p,q}^n(\mathbb{R};R)$ equipped with the radius $R>0$, which controls the distance between our proposed threshold $\rho_{i,\kappa}$ and the concerned one corresponding to $f_X(\lambda)$. Following from \cite[Theorem 8]{donoho1994}, for $f_X(\lambda) \in \mathcal{B}_{p,q}^n(\mathbb{R};R)$, there is the optimal rate of  mean-square convergence, that is, $T^{-2n/(2n+1)}$. Based on the results above, from \citep[Theorem 5.1 and Proposition 3.1]{neumann1996}, it is observed that there exists some thresholds $\rho_{i,\kappa}$ in which 
\[
\rho_{i,\kappa} = \Theta\left( T^{-1/2} \right) \cdot \left( \int_{-\pi}^{\pi} \varphi_{i,\kappa}(\lambda) f_X(\lambda) \dif \lambda \right) \cdot \sqrt{2 \log(|\mathfrak{J}_T|_{\#}) } ~,
\]
such that 
\[
\sup_{f_X \in \mathcal{B}_{p,q}^n(\mathbb{R};R)} \left\{ \mathbb{E} \left[ \left\| \mathbf{J}_T(\lambda_j) - f_X(\lambda_j) \right\|_{L_2([-\pi,\pi])} \right] \right\} = \mathcal{O} \left( (\log T/T)^{2n/(2n+1)} \right) .
\]

\noindent\textbf{Consistency of $\widehat{G}_{\textrm{ASE}}$.} According to the near-optimal rate bound above, we can derive that $T^{\beta}( \mathbf{J}_T(\lambda) - f_X(\lambda) ) \stackrel{\mathrm{P}}{\longrightarrow} 0$ for $\beta \in (0,1)$ and $\boldsymbol{d}^0 \in \Omega_{\beta}$. For convenience, we abbreviate it as $\mathbf{J}_T(\lambda) = f_X(\lambda) + \op(T^{-\beta})$ for $\beta \in (0,1)$ and $\boldsymbol{d}^0 \in \Omega_{\beta}$. We unfold $\widehat{G}_{\textrm{ASE}}$ as
\[
\begin{aligned}
	\widehat{G}_{\textrm{ASE}} (\boldsymbol{d}^0) &= \frac{1}{m} \sum_{j=1}^{m} \re\left[\Psi_{j}(\boldsymbol{d})^{-1} \mathbf{J}_T(\lambda_j) \overline{\Psi_{j}(\boldsymbol{d})}^{-1}\right] \\
	&= \frac{1}{m} \sum_{j=1}^{m} \re\left[\Psi_{j}(\boldsymbol{d})^{-1} \left( f_X(\lambda_j) + \op(T^{-\beta}) \right) \overline{\Psi_{j}(\boldsymbol{d})}^{-1}\right] \\
	&= G^0 + \frac{1}{m} \sum_{j=1}^{m} \re\left[\Psi_{j}(\boldsymbol{d})^{-1} \op(T^{-\beta}) \overline{\Psi_{j}(\boldsymbol{d})}^{-1}\right] \\
	&= G^0 + \frac{1}{m} \sum_{j=1}^{m} \re\left[\Psi_{j}(\boldsymbol{d})^{-1} \overline{\Psi_{j}(\boldsymbol{d})}^{-1}\right] \op(T^{-\beta})  .
\end{aligned}
\]
Thus, for $p,q \in [l]$, we have
\begin{equation} \label{eq:use2}
	\begin{aligned}
		\widehat{G}_{\textrm{ASE}} (\boldsymbol{d}^0)_{p,q} &= G^0_{p,q} + \frac{1}{m} \sum_{j=1}^{m} \re\left[ \e^{\ii (\pi-\lambda) (d_p^0-d_q^0)/2} \lambda^{-d_p^0-d_q^0} \right] \op(T^{-\beta}) \\
		&= G^0_{p,q} + \left[ \frac{1}{m} \sum_{j=1}^m \lambda_j^{-d^0_p-d^0_q} \right] \op(T^{-\beta}) \\
		&= G^0_{p,q} + \frac{1}{d_p^0+d_q^0+1} \left[ \frac{d_p^0+d_q^0+1}{m} \sum_{j=1}^m \binom{j}{m}^{d_p^0+d_q^0} \right] \left(\frac{2\pi m}{T}\right)^{d_p^0+d_q^0} \op(T^{-\beta}) \\
		&= G^0_{p,q} + \frac{1}{d_p^0+d_q^0+1} \left[ \mathcal{O}\left( \frac{1}{m^{1-\beta}} \right) + 1 \right] \left(\frac{2\pi m}{T}\right)^{d_p^0+d_q^0} \op(T^{-\beta}) \\
		&= G^0_{p,q} + \op(1),
	\end{aligned} 
\end{equation}
where the fourth equality follows from Lemma~\ref{tec:robinson} by taking $\gamma = d_p^0+d_q^0+1 > 1- \beta = \epsilon$. Hence, it holds
\[
\widehat{G}_{\textrm{ASE}} (\boldsymbol{d}^0)_{p,q} = G^0_{p,q} + \op(1) , 
\]
for $p,q \in [l]$, $\beta \in (0,1)$, and $\boldsymbol{d}^0 \in \Omega_{\beta}$. This proof is complete. $\hfill\square$

\section{Assumptions of Theorem~\ref{thm:consistency-an}}
\label{app:assumptions}
Now, we list the assumptions we need, which are exactly the same as those of GSE~\cite{shimotsu2007:GSE}.
\begin{assumption} \label{ass:1}
	For $p,q\in[l]$, let $f_{pq}(\lambda)$ and $G_{pq}^0$ indicate the $(p,q)$-th element of $f_X(\lambda)$ and $G_0$, respectively. As $\lambda \to 0^+$, we have
	\[
	f_{pq}(\lambda) = \e^{\ii (\pi-\lambda) (d_p^0-d_q^0)/2} G_{pq}^0 \lambda^{-d_p^0-d_q^0} + \mathrm{o}\left( \lambda^{-d_p^0-d_q^0} \right).
	\]
\end{assumption}
Assumption~\ref{ass:1} describes the true spectral density matrix behavior at the origin.
\begin{assumption} \label{ass:2}
	The concerned process $\{\mathbf{X}_t\}$ leads to a causal system, i.e.,
	\begin{equation} \label{eq:A}
		\mathbf{X}_t = A(\mathfrak{B})\boldsymbol{\epsilon}_t = \sum_{j=0}^{\infty} A_j\boldsymbol{\epsilon}_{t-j} ~~\text{with}~~ \sum_{j=0}^{\infty}\|A_j\|_{\infty}^2 < \infty ,
	\end{equation}
	where for $t \in\mathbb{Z}$, it holds
	\[
	\mathbb{E}\left(\boldsymbol{\epsilon}_t \!\mid\! \mathscr{F}_{t-1}\right)=0 \quad\text{and}\quad \mathbb{E}\left( \boldsymbol{\epsilon}_{t} \boldsymbol{\epsilon}_t^{\top} \!\mid\! \mathscr{F}_{t-1} \right) = \mathbf{1}_{l \times l} ,
	\]
	where $\mathscr{F}_t$ denotes the $\sigma$-field generated by $\{ \boldsymbol{\epsilon}_s, s \leq t \}$, and there exist a scalar random variable $\xi$ and a constant $K > 0$ such that $\mathbb{E}(\xi^2) < \infty$ and
	\[
	\mathbb{P}\left(\left\|\boldsymbol{\epsilon}_t\right\|_{\infty}^{2}>\eta\right) \leq K \mathbb{P}\left(\xi^{2}>\eta\right), \quad\text{for}\quad \forall \eta>0.
	\]
\end{assumption}
Assumption~\ref{ass:2} states the causality of $\mathbf{X}_t$, where the behavior of the source process which is assumed to be a not necessarily uncorrelated square integrable martingale difference uniformly dominated (in probability) by a scalar random variable with finite second moment.
\begin{assumption} \label{ass:3}
	In a neighborhood $(0,\delta)$ of the origin, $A$ given by Eq.~\eqref{eq:A} is differentiable and satisfies
	\[
	{\partial A_j(\lambda)} / {\partial \lambda} = \mathcal{O} \left( \lambda^{-1} \left\| A_j(\lambda) \right\|\right) \quad\text{as}\quad \lambda \to 0+ .
	\]
\end{assumption}
Assumption~\ref{ass:3} provides a regularity condition, which implies
\[
{\partial A_j(\lambda)} / {\partial \lambda} = \mathcal{O}\left( \lambda^{-d_j-1}\right)
\]
since $\| A_j(\lambda) \| \leq (A_j(\lambda)\overline{A_j(\lambda)})^{1/2} = \sqrt{2\pi f_{jj}(\lambda)} $.
\begin{assumption} \label{ass:4}
	As $T \to \infty$, we have
	\[
	1/m + m/T \to 0.
	\]
\end{assumption}
Assumption~\ref{ass:4} is minimal but necessary since $m=|\{\lambda_j\}|_{\#}$ must go to infinity for consistency, but slower than the number of observed timestamps $T$ in view of Assumption~\ref{ass:1}. Notice that Assumptions~\ref{ass:1}-\ref{ass:4} are only concerned to the behavior of the spectral density matrix on a neighborhood of the origin and no assumption whatsoever is made on the spectral density matrix behavior outside this neighborhood.

\begin{assumption} \label{ass:B1}
	For $\alpha \in (0,2]$ and $p,q\in[l]$, as $\lambda \to 0^+$,
	\[
	f_{pq}(\lambda) = \e^{\ii (\pi-\lambda) (d_p^0-d_q^0)/2} G_{pq}^0 \lambda^{-d_p^0-d_q^0} + \mathrm{o}\left( \lambda^{-d_p^0-d_q^0+\alpha} \right).
	\]
\end{assumption}
Assumption~\ref{ass:B1} is a smoothness condition often imposed in spectral analysis.
\begin{assumption} \label{ass:B2}
	Assumption~\ref{ass:2} holds and $\mathbb{E}(\boldsymbol{\epsilon}_t^4) < \infty$.
\end{assumption}

\begin{assumption} \label{ass:B3}
	For $\alpha \in (0,2]$ and any $\delta>0$,
	\[
	\frac{1}{m}+\frac{m^{1+2 \alpha} (\log m)^2}{T^{2 \alpha}}+\frac{\log T}{m^{\delta}} \to 0 \quad\text{as}\quad T \to \infty .
	\]
\end{assumption}
Assumption~\ref{ass:B3} provides a necessary numerical condition, which implies that
\[
(m / T)^{c} = \mathfrak{o} \left( m^{-\frac{c}{2 \alpha}} (\log m)^{-\frac{c}{\alpha}} \right), \quad\text{for}\quad c \neq 0 .
\]
\begin{assumption} \label{ass:B4}
	There exists $\mathbf{M} \in \mathbb{R}^{l \times k}$ such that
	\[
	\Psi_j \left( \boldsymbol{d}^0 \right)^{-1} A(\lambda_j)= \mathbf{M} + \mathrm{o}(1) \quad\text{as}\quad \lambda_j \to 0 ,
	\]
	which implies $\mathbf{M} \mathbf{M}^{\top} = 2\pi G^0$.
\end{assumption}
Assumption~\ref{ass:B4} is a mild regularity condition in the approximation degree of $A(\lambda_j)$ by $\Psi_j(\boldsymbol{d}^0)$.

\section{Full Proofs for Theorem~\ref{thm:consistency-an}} 
\label{app:consistency-an}

The proof idea of consistency is similar to those provided by Lobato~\citep{lobato1999:two} and Shimotsu~\citep{shimotsu2007:GSE}. Given
\[
S(\boldsymbol{d}) = \left( \log \big| \widehat{G}_\textrm{ASE}(\boldsymbol{d}) \big| - \frac{2}{m} \sum_{i=1}^{l} \sum_{j=1}^{m} d_i \log \lambda_j \right) - \left( \log \big| \widehat{G}_\textrm{ASE}(\boldsymbol{d}^0) \big| - \frac{2}{m} \sum_{i=1}^{l} \sum_{j=1}^{m} d_i^0 \log \lambda_j \right) 
\]
and
\[
\left\{\begin{aligned}
	& \boldsymbol{d}_{\mathrm{ASE}} - \boldsymbol{d}^0 = \Delta\boldsymbol{d} \overset{\underset{\mathrm{def}}{}}{=} (d'_1, \dots, d'_l) ,\\
	& \Omega_{\delta} = \{ \boldsymbol{d} \mid \| \boldsymbol{d} - \boldsymbol{d}^0 \|_{\infty} > \delta \}, \quad\text{for}\quad 0<\delta<1/2 ,\\
	& \Omega_1 = \{ \boldsymbol{d} \mid \boldsymbol{d} \in [-1/2 + \epsilon, 1/2]^l \}, \quad\text{for}\quad 0<\epsilon<1/4 ,\\
	& \Omega_2 = \Omega_{\beta} \setminus \Omega_1 ,\\
\end{aligned} \right.
\]
we have
\[
\begin{aligned}
	\mathbb{P} \left(\left\| \widehat{\boldsymbol{d}}_{\mathrm{ASE}} - \boldsymbol{d}_{0} \right\|_{\infty} > \delta\right) &\leq \mathbb{P} \left( \inf_{\overline{\Omega_{\delta}} \cap \Omega_{\beta}} S(\boldsymbol{d})  \leq 0 \right) \\
	&\leq \mathbb{P} \left( \inf_{\overline{\Omega_{\delta}} \cap \Omega_1} S(\boldsymbol{d}) \leq 0 \right) + \mathbb{P} \left( \inf_{\Omega_2} S(\boldsymbol{d})  \leq 0 \right) \\ &\overset{\underset{\mathrm{def}}{}}{=} P_1 + P_2 .
\end{aligned}
\]
It's sufficient to show $P_1 \to 0$ and $P_2 \to 0$ as $T \to \infty$.

\noindent\textbf{About $\boldsymbol{P_1}$.} Unfold $S(\boldsymbol{d})$ as 
\[
\begin{aligned}
	S(\boldsymbol{d}) =&~ \log \left| \widehat{G}_{\mathrm{ASE}}(\boldsymbol{d}) \right| - \log \left| \widehat{G}_{\mathrm{ASE}}( \boldsymbol{d}^0 ) \right| - \frac{2}{m} \sum_{i=1}^{l} \sum_{j=1}^{m} d'_i \log \lambda_j \\
	=&~ \log \left| \widehat{G}_{\mathrm{ASE}}(\boldsymbol{d}) \right| - \log \left| \widehat{G}_{\mathrm{ASE}} ( \boldsymbol{d}^0 ) \right| + \log \left( \frac{2 \pi m}{T} \right)^{-2 \sum_i d'_i } \\
	&~- 2 \sum_{i=1}^{l} d'_i \left( \frac{1}{m} \sum_{j=1}^{m} \log j - \log m \right) - \sum_{i=1}^{l} \log(2 d'_i + 1) . \\
\end{aligned}
\]
Define
\[
\left\{\begin{aligned}
	& A(\boldsymbol{d}) = \left( \frac{2 \pi m}{T} \right)^{-2 \sum_i d'_i } \left| \widehat{G}_{\mathrm{ASE}}(\boldsymbol{d}) \right| ,\\
	& B(\boldsymbol{d}) = \prod_{i=1}^{l} \frac{1}{2d'_i + 1} \left| \widehat{G}^0 \right| ,\\
	& S_r(\boldsymbol{d}) = - 2 \sum_{i=1}^{l} d'_i \left( \frac{1}{m} \sum_{j=1}^{m} \log j - \log m \right) - \sum_{i=1}^{l} \log(2 d'_i + 1) .
\end{aligned} \right.
\]
Thus, $S(\boldsymbol{d})$ can be rewritten as
\[
\begin{aligned}
	S(\boldsymbol{d}) &= \log A(\boldsymbol{d}) - \log B(\boldsymbol{d}) - \log A(\boldsymbol{d}^0) + \log B(\boldsymbol{d}^0) + S_r(\boldsymbol{d}) \\
	&= \left[ \log A(\boldsymbol{d}) - \log B(\boldsymbol{d}) \right] - \left[ \log A(\boldsymbol{d}^0) - \log B(\boldsymbol{d}^0) \right] + S_r(\boldsymbol{d}) .
\end{aligned}
\]
For $A(\boldsymbol{d})$ and $A(\boldsymbol{d})$, it suffices to show that there exits a function $H(\boldsymbol{d})$ satisfying
\[
\text{(i)}~ \sup_{\Omega_1} \left| A(\boldsymbol{d}) - H(\boldsymbol{d}) \right| = \op(1) ,
\quad
\text{(ii)}~ H(\boldsymbol{d}) \geq B(\boldsymbol{d}) ,
\quad
\text{(iii)}~ H(\boldsymbol{d}^0) = B(\boldsymbol{d}^0) ,
\quad\text{as}\quad T \to \infty .
\] 
To prove (i), recall that
\[
\begin{aligned}
	\Psi_j(\boldsymbol{d})^{-1} &= \diag_{i \in\{1, \cdots, l\}} \{ \lambda_j^{d_i} \e^{\ii (\lambda_j-\pi)d_i/2} \} \\
	&= \diag_{i \in\{1, \cdots, l\}} \{ \lambda_j^{(d_i-d^0_i)} \e^{\ii (\lambda_j-\pi)(d_i-d^0_i)/2} ~ \lambda_j^{d^0_i} \e^{\ii (\lambda_j-\pi) d^0_i/2} \} \\
	&= \Psi_j(\boldsymbol{d}-\boldsymbol{d}^0)^{-1} \Psi_j(\boldsymbol{d}^0)^{-1} \\
	&= \Psi_j(\Delta\boldsymbol{d})^{-1} \Psi_j(\boldsymbol{d}^0)^{-1} ,
\end{aligned}
\]
and then, we have
\[
\resizebox{1\hsize}{!}{$
	\begin{aligned}
		A(\boldsymbol{d}) &= \left( \frac{2 \pi m}{T} \right)^{-2 \sum_i d'_i } \left| \widehat{G}_{\mathrm{ASE}}(\boldsymbol{d}) \right| \\
		&= \left( \frac{2 \pi m}{T} \right)^{-2 \sum_i d'_i } \left| \frac{1}{m} \sum_{j=1}^{m} \re\left[\Psi_j(\boldsymbol{d})^{-1} \mathbf{J}_T(\lambda_j) \overline{\Psi_j(\boldsymbol{d})^{-1}}\right] \right| \\
		&= \left( \frac{2 \pi m}{T} \right)^{-2 \sum_i d'_i } \left| \frac{1}{m} \sum_{j=1}^{m} \re\left[\Psi_j(\Delta\boldsymbol{d})^{-1} \Psi_j(\boldsymbol{d}^0)^{-1} \mathbf{J}_T(\lambda_j) \overline{\Psi_j(\boldsymbol{d}^0)^{-1}}~ \overline{\Psi_j(\Delta\boldsymbol{d})^{-1}} \right] \right| \\
		&= \left( \frac{2 \pi m}{T} \right)^{-2 \sum_i d'_i } \left| \frac{1}{m} \sum_{j=1}^{m} \re\left[\Psi_j(\Delta\boldsymbol{d})^{-1} \Psi_j(\boldsymbol{d}^0)^{-1} \left( f_X(\lambda_j) + \op(T^{-\beta}) \right) \overline{\Psi_j(\boldsymbol{d}^0)^{-1}}~ \overline{\Psi_j(\Delta\boldsymbol{d})^{-1}} \right] \right| \\
		&= \left| \frac{1}{m} \sum_{j=1}^{m} \re\left[ M_j(\Delta\boldsymbol{d}) G^0  \overline{M_j(\Delta\boldsymbol{d})} \right] + \frac{1}{m} \sum_{j=1}^{m} \re\left[ M_j(\Delta\boldsymbol{d}) \Psi(\boldsymbol{d}^0)^{-1} \op(T^{-\beta}) \overline{\Psi(\boldsymbol{d}^0)^{-1}}~ \overline{M_j(\Delta\boldsymbol{d})} \right] \right| ,
	\end{aligned} $}
\]
where $M_j(\Delta\boldsymbol{d}) = \diag_{i \in\{1, \cdots, l\}} \left\{ \e^{\ii (\lambda_j-\pi)d'_i/2}(j/m)^{d'_i} \right\}$ . Similar to Eq.~\eqref{eq:use2}, for $p,q \in [l]$, we have
\[
\begin{aligned}
	&\frac{1}{m} \sum_{j=1}^{m} \re\left[ M_j(\Delta\boldsymbol{d}) \Psi_j(\boldsymbol{d}^0)^{-1} \op(T^{-\beta}) \overline{\Psi_j(\boldsymbol{d}^0)^{-1}}~ \overline{M_j(\Delta\boldsymbol{d})} \right]_{p,q} \\
	&\quad= \frac{1}{m} \sum_{j=1}^{m} \re\left[ \e^{\ii (\lambda_j-\pi) (d_p-d_q)/2} \lambda_j^{d_p^0+d_q^0} \left(j/m\right)^{d'_p+d'_q} \right] \op(T^{-\beta}) \\
	&\quad\leq \frac{C}{m} \sum_{j=1}^m \left( \frac{j}{m} \right)^{d_p^0+d_q^0} \left( \frac{m}{T} \right)^{d_p^0+d_q^0} \op(T^{-\beta}) 
	\\
	&\quad= \frac{1}{d_p+d_q+1} \left(\frac{2\pi m}{T}\right)^{d_p+d_q} \left[ \mathcal{O}\left( \frac{1}{m^{1-\beta}} \right) + 1 \right] \left( \frac{m}{T} \right)^{d_p^0+d_q^0} \op(T^{-\beta}) \\
	&\quad= \op(1),
\end{aligned} 
\]
where $C$ is a constant, the first equality holds according to $d_p = d'_p + d^0_p$ for $p \in [m]$, and the inequality follows from Lemma~\ref{tec:robinson}. Hence, for $\boldsymbol{d} \in \Omega_{\beta}$ and $\Delta\boldsymbol{d} \in \Omega_1$, we have
\[
A(\boldsymbol{d}) = \left| \frac{1}{m} \sum_{j=1}^{m} \re\left[ M_j(\Delta\boldsymbol{d}) G^0 \overline{M_j(\Delta\boldsymbol{d})} \right] + \op(1) \right| .
\]
To construct the objective function $H$ in (i), we should proceed to derive an approximation of the right hand side. From Lemma~\ref{tec:robinson}, we have
\[
\sup _{C \geq \gamma \geq \varepsilon} \left| \frac{\gamma}{m} \sum_{j=1}^{m} \left(\frac{j}{m}\right)^{\gamma-1}-1 \right| = \mathcal{O} \left( \frac{1}{m^{\varepsilon}} \right) \quad\text{as}\quad T \to \infty, \quad\text{for}\quad \epsilon \in (0,1) \quad\text{and}\quad C \in (\epsilon, \infty) .
\]
Further, we have $ \e^{\ii (\lambda - \pi) (d'_p - d'q)/2} = \e^{\ii \pi (d'_p - d'q)/2} + \mathcal{O}(\lambda) $. Define two matrices $\mathbf{U}(\Delta\boldsymbol{d})$ and $\mathbf{V}(\Delta\boldsymbol{d})$
\[
\mathbf{U}_{p,q}(\Delta\boldsymbol{d}) = \e^{\ii \pi (d'_p - d'q)/2}
\quad\text{and}\quad
\mathbf{V}_{p,q}(\Delta\boldsymbol{d}) = \frac{1}{d'_p+d'_q+1} = \int_{0}^{1} x^{d'_p+d'_q} \dif x ,
\quad\text{for}\quad p,q \in[l] .
\]
It's observed that
\[
\frac{1}{m} \sum_{j=1}^{m} \left[ M_j(\Delta\boldsymbol{d}) G^0 \overline{M_j(\Delta\boldsymbol{d})} \right] = \mathbf{U}(\Delta\boldsymbol{d}) \odot G^0 \odot \mathbf{V}(\Delta\boldsymbol{d}) + \mathcal{O}\left( \frac{m}{T} \right) + \frac{1}{m^{2\delta}} ,
\]
where $\odot$ denotes the Hadamard product. Thus, the function $H(\boldsymbol{d}) \overset{\underset{\mathrm{def}}{}}{=} \left| \re[\mathbf{U}(\Delta\boldsymbol{d})] \odot G^0 \odot \mathbf{V}(\Delta\boldsymbol{d}) \right|$ is an alternative solution for (i).

To prove (ii), we have
\[
\log A(\boldsymbol{d}) - \log B(\boldsymbol{d}) \geq \log A(\boldsymbol{d}) - \log H(\boldsymbol{d}) = \log (H(\boldsymbol{d}) + \op(1)) - \log H(\boldsymbol{d}) = \op(1),
\]
for $\boldsymbol{d} \in \Omega_{\beta}$ and $\Delta\boldsymbol{d} \in \Omega_1$, since $\inf_{\Omega_1} H(\boldsymbol{d}) > \inf_{\Omega_1} H(\boldsymbol{d}) > 0$.

To prove (iii), similarly, we have
\[
\log A(\boldsymbol{d}^0) - \log B(\boldsymbol{d}^0) = \log (H(\boldsymbol{d}^0) + \op(1)) - \log H(\boldsymbol{d}^0) = \op(1) .
\]
Therefore, the concerned function $H(\boldsymbol{d})$ satisfies (i), (ii), and (iii) above as desired.

For $S_r(\boldsymbol{d})$, we have (from Lemma~\ref{tec:robinson:2})
\[
\log m - \frac{1}{m} \sum_{j=1}^{m} \log j = 1 + \mathcal{O}\left(\frac{\log m}{m}\right).
\]
Thus, we have
\[
S_r(\boldsymbol{d}) = \sum_{i=1}^{l} \left[ 2d'_i - \log(2d'_i + 1) \right] + \mathcal{O}\left(\frac{\log m}{m}\right).
\]
It's observed that $x - \log(x+1)$ has a unique minimum in $(-1,+\infty)$ at $x=0$ and $x- \log(x+1) \geq x^2/4 $ in $(-1,1]$. Thus, provided $x = 2d'_i$, one has
\[
\inf_{\overline{\Omega_{\delta}} \cap \Omega_1} S_r(\boldsymbol{d}) \geq \delta^2 >0 .
\]
Finally, we can conclude
\[
P_1 = \mathbb{P} \left( \inf_{\overline{\Omega_{\delta}} \cap \Omega_1} S(\boldsymbol{d}) \leq 0 \right) \to 0 \quad\text{as}\quad T \to \infty.
\]

\noindent\textbf{About $\boldsymbol{P_2}$.} Re-define
\[
\left\{\begin{aligned}
	A(\boldsymbol{d}) &= \frac{1}{m} \sum_{j=1}^{m} \re\left[ B_j(\Delta\boldsymbol{d})  \Psi_j(\boldsymbol{d}^0)^{-1} \mathbf{J}_T(\lambda_j) \overline{\Psi_j(\boldsymbol{d}^0)^{-1}}~ \overline{B_j(\Delta\boldsymbol{d})^{-1}} \right]  ,\\
	B_j(\boldsymbol{d}) &= \diag_{i \in\{1, \cdots, l\}} \left\{ \e^{\ii (\lambda_j - \pi)d'_i/2 } (j/K)^{d'_i} \right\} ,\\
	K &\overset{\underset{\mathrm{def}}{}}{=} \exp\left( \frac{1}{m} \sum_{j=1}^m \log j  \right) .
\end{aligned} \right.
\]
Then we can rewrite $S(\boldsymbol{d})$ as 
\[
S(\boldsymbol{d}) = \log \left| \widehat{G}_{\mathrm{ASE}}(\boldsymbol{d}) \right| - \log \left| \widehat{G}_{\mathrm{ASE}}( \boldsymbol{d}^0 ) \right| - \frac{2}{m} \sum_{i=1}^{l} \sum_{j=1}^{m} d'_i \log \lambda_j = \log \left| A(\boldsymbol{d}) \right| - \log \left| A(\boldsymbol{d}^0) \right| .
\]
To prove that $P_2 = \mathbb{P} \left( \inf_{\Omega_2} S(\boldsymbol{d}) \leq 0 \right) \to 0$ as $T \to \infty$, it suffices to show
\[
\mathbb{P} \left( \inf_{\Omega_2} \left| A(\boldsymbol{d}) \right| - \left| A(\boldsymbol{d}^0) \right| \leq 0 \right) \to 0 \quad\text{as}\quad T \to \infty,
\]
since $\log$ is a monotone increasing function. Observe that $K \sim m / \e$ as $m \to \infty $ and $A(\boldsymbol{d})$ is a sum of $m$ positive semi-definite matrices since
\[
\re\left[ B_j(\Delta\boldsymbol{d}) \Psi_j(\boldsymbol{d}^0)^{-1} \mathbf{J}_T(\lambda_j) \overline{\Psi_j(\boldsymbol{d}^0)^{-1}}~ \overline{B_j(\Delta\boldsymbol{d})^{-1}} \right] = \re\left[ \boldsymbol{\mu_j} \boldsymbol{\mu_j}^{\top} \right] ,
\]
for some apposite $l$-vector $\boldsymbol{\mu_j}$. Then, for $\kappa \in (0,1)$, we can define
\[
\left\{\begin{aligned}
	A_{\kappa}(\boldsymbol{d}) &= \frac{1}{m} \sum_{j=\lfloor m\kappa \rfloor}^{m} \re\left[ B_j(\Delta\boldsymbol{d})  \Psi_j(\boldsymbol{d}^0)^{-1} \mathbf{J}_T(\lambda_j) \overline{\Psi_j(\boldsymbol{d}^0)^{-1}}~ \overline{B_j(\Delta\boldsymbol{d})^{-1}} \right]  ,\\
	F_{\kappa}(\boldsymbol{d}) &= \frac{1}{m} \sum_{j=\lfloor m\kappa \rfloor}^{m} \re\left[ B_j(\Delta\boldsymbol{d}) G^0 \overline{B_j(\Delta\boldsymbol{d})^{-1}} \right] ,\\
\end{aligned} \right.
\]
and thus, one has
\[
\begin{aligned}
	A_{\kappa}(\boldsymbol{d}) &= \frac{1}{m} \sum_{j=\lfloor \kappa m \rfloor}^{m} \re\left[ B_j(\Delta\boldsymbol{d}) \Psi_j(\boldsymbol{d}^0)^{-1} \left( f_X(\lambda_j) + \op(T^{-\beta}) \right) \overline{\Psi_j(\boldsymbol{d}^0)^{-1}}~ \overline{B_j(\Delta\boldsymbol{d})^{-1}} \right] \\
	&= F_{\kappa}(\boldsymbol{d}) + \frac{1}{m} \sum_{j=\lfloor \kappa m \rfloor}^{m} \re\left[ B_j(\Delta\boldsymbol{d}) \Psi_j(\boldsymbol{d}^0)^{-1} \op(T^{-\beta}) \overline{\Psi_j(\boldsymbol{d}^0)^{-1}}~ \overline{B_j(\Delta\boldsymbol{d})^{-1}} \right]
\end{aligned}
\]
Similar to Eq.~\eqref{eq:use2}, for $p,q \in [l]$, we have
\[
\begin{aligned}
	&\frac{1}{m} \sum_{j=\lfloor \kappa m \rfloor}^{m} \re\left[ B_j(\Delta\boldsymbol{d}) \Psi_j(\boldsymbol{d}^0)^{-1} \op(T^{-\beta}) \overline{\Psi_j(\boldsymbol{d}^0)^{-1}}~ \overline{B_j(\Delta\boldsymbol{d})^{-1}} \right]_{p,q} \\
	&\quad= \frac{1}{m} \sum_{j=\lfloor \kappa m\rfloor}^{m} \re\left[ \e^{\ii (\lambda_j-\pi) (d_p-d_q)/2} \left(\frac{2\pi j}{T}\right)^{d_p^0+d_q^0} \left(\frac{j}{K}\right)^{d'_p+d'_q} \right] \op(T^{-\beta}) \\
	&\quad= \mathcal{O}(1)~ \frac{1}{m} \sum_{j=\lfloor \kappa m \rfloor}^m \left( \frac{j}{m} \right)^{-2(d'_p+d'_q)} \left( \frac{m}{K} \right)^{d_p+d_q} \left( \frac{m}{T} \right)^{d_p^0+d_q^0} \op(T^{-\beta})  
	\\
	&\quad= \mathcal{O}(1)~ \left( \frac{m}{K} \right)^{d_p+d_q} \op(T^{-\beta}) \\
	&\quad= \op(1),
\end{aligned} 
\]
where the third equality follows from Lemma~\ref{tec:shimotsu2005}. Hence, for  for $\boldsymbol{d} \in \Omega_{\beta}$ and $\Delta\boldsymbol{d} \in \Omega_2$, we have
\[
\sup_{\Omega_2} \left\{ \Big| |A_{\kappa}(\boldsymbol{d})| - |F_{\kappa}(\boldsymbol{d})| \Big| \right\} = \op(1) \quad\text{as}\quad T \to \infty .
\]
From \citep[Theorem 1]{shimotsu2007:GSE}, we can conclude
\[
P_2 = \mathbb{P} \left( \inf_{\Omega_2} S(\boldsymbol{d}) \leq 0 \right) \to 0 \quad\text{as}\quad T \to \infty .
\]
In summary, we finish the consistency proof of Theorem~\ref{thm:consistency-an}.

The proof idea of asymptotic normality is similar to those in \citep{lobato1999:two} and \citep{shimotsu2007:GSE}. Based on the consistency analysis of $\widehat{\boldsymbol{d}}_{\textrm{ASE}}$ above, the following holds with probability tending to one,
\[
\begin{aligned}
	0 &= \left.\frac{\dif }{\dif \boldsymbol{d} }\right|_{\widehat{\boldsymbol{d}}_{\textrm{ASE}}} \left\{ \log \big| \widehat{G}_\textrm{ASE}(\boldsymbol{d}) \big| - \frac{2}{m} \sum_{i=1}^{l} \sum_{j=1}^{m} d_i \log \lambda_j \right\} \\
	&= \Bigg[ \left.\frac{\dif }{\dif \boldsymbol{d}}\right|_{\boldsymbol{d}^0} + \left( \widehat{\boldsymbol{d}}_{\mathrm{ASE}} - \boldsymbol{d}^0 \right) \left.\frac{\dif^{2} }{\dif \boldsymbol{d}^{\top} \dif \boldsymbol{d} }\right|_{\bar{\boldsymbol{d}}} \Bigg] \left( \log \big| \widehat{G}_\textrm{ASE}(\boldsymbol{d}) \big| - \frac{2}{m} \sum_{i=1}^{l} \sum_{j=1}^{m} d_i \log \lambda_j \right) ,
\end{aligned}
\]
for some $\bar{\boldsymbol{d}}$ such that $\| \bar{\boldsymbol{d}} - \boldsymbol{d}^0 \|_{\infty} \leq \| \widehat{\boldsymbol{d}}_{\textrm{ASE}} -\boldsymbol{d}^0 \|_{\infty}$, as $T$ goes to infinity. It's observed that $\widehat{\boldsymbol{d}}_{\textrm{ASE}}$ has the stated limiting distribution if the followings hold
\begin{equation} \label{eq:an_1}
	\left.\sqrt{m}~ \frac{\partial }{\partial \boldsymbol{d}}\right|_{\boldsymbol{d}^0} \left( \log \big| \widehat{G}_\textrm{ASE}(\boldsymbol{d}) \big| - \frac{2}{m} \sum_{i=1}^{l} \sum_{j=1}^{m} d_i \log \lambda_j \right)  \stackrel{\mathrm{d}}{\longrightarrow} \mathcal{N}(0, \Sigma^{-1})
\end{equation}
and 
\begin{equation} \label{eq:an_2}
	\left. \frac{\partial^{2} }{\partial \boldsymbol{d}^{\top} \partial \boldsymbol{d} } \right|_{\bar{\boldsymbol{d}}} \left( \log \big| \widehat{G}_\textrm{ASE}(\boldsymbol{d}) \big| - \frac{2}{m} \sum_{i=1}^{l} \sum_{j=1}^{m} d_i \log \lambda_j \right) \stackrel{\mathrm{P}}{\longrightarrow} \Sigma^{-1} ,
\end{equation}
as $T \to \infty$, where
\[
\Sigma = \frac{4+\pi^2}{2} G^0 \odot\left(G^0\right)^{-1} + \frac{4-\pi^{2}}{2} \mathbf{1}_{l \times l}  .
\]

\noindent\textbf{Score vector approximation.} Observe that, for $i \in [l]$,
\[
\resizebox{1\hsize}{!}{$
	\sqrt{m}~ \frac{\partial }{\partial d_i} \left( \log \big| \widehat{G}_\textrm{ASE}(\boldsymbol{d}) \big| - \frac{2}{m} \sum_{i=1}^{l} \sum_{j=1}^{m} d_i \log \lambda_j \right) = \sqrt{m} \tr\left[ \widehat{G}_{\mathrm{ASE}}(\boldsymbol{d})^{-1} \frac{\partial \widehat{G}_\textrm{ASE}(\boldsymbol{d})}{\partial d_i} \right] - \frac{2}{\sqrt{m}} \sum_{j=1}^{m} \log \lambda_j .
	$}
\]
Unfolding the first term of the right hand above, we have 
\[
\begin{aligned}
	\left. \sqrt{m}~ \frac{\partial \widehat{G}_\textrm{ASE}(\boldsymbol{d})}{\partial d_i} \right|_{\boldsymbol{d}^0} =&~ \frac{1}{\sqrt{m}} \sum_{j=1}^{m} \re\left[ \left(\log\lambda_j+\ii\frac{\pi-\lambda_j}{2}\right) \Psi_j(\boldsymbol{d}^0)^{-1} \mathbf{J}_{:,i}(\lambda_j) \overline{\Psi_j(\boldsymbol{d}^0)^{-1}} \right] \\
	&~+ \frac{1}{\sqrt{m}} \sum_{j=1}^{m} \re\left[ \left(\log\lambda_j+\ii\frac{\lambda_j-\pi}{2}\right) \Psi_j(\boldsymbol{d}^0)^{-1} \mathbf{J}_{i,:}(\lambda_j) \overline{\Psi_j(\boldsymbol{d}^0)^{-1}} \right] \\
	=&~ \frac{1}{\sqrt{m}} \sum_{j=1}^{m} \log \lambda_j \re\left[ \Psi_j(\boldsymbol{d}^0)^{-1} \left( \mathbf{J}_{:,i}(\lambda_j) + \mathbf{J}_{i,:}(\lambda_j) \right) \overline{\Psi_j(\boldsymbol{d}^0)^{-1}} \right] \\
	&~+ \frac{1}{\sqrt{m}} \sum_{j=1}^{m} \frac{\lambda_j-\pi}{2} \im\left[ \Psi_j(\boldsymbol{d}^0)^{-1} \left( -\mathbf{J}_{:,i}(\lambda_j) + \mathbf{J}_{i,:}(\lambda_j) \right) \overline{\Psi_j(\boldsymbol{d}^0)^{-1}} \right] ,
\end{aligned}
\]
where $\mathbf{J}_{:,i}(\lambda_j)$ and $\mathbf{J}_{i,:}(\lambda_j)$ are two $l \times l$ matrices whose $i$-th column and $i$-th row vectors are the same as those of $\mathbf{J}_T(\lambda_j)$, respectively, and all other elements are zero. Thus, for any vector $\boldsymbol{\eta} = (\eta_1, \dots, \eta_l) \in \mathbb{R}^l$, we have
\begin{equation} \label{eq:use1}
	\sqrt{m}~ \sum_{i=1}^{l} \eta_i \left. \frac{\partial}{\partial d_i} \right|_{\boldsymbol{d}^0} \left\{ \log \big| \widehat{G}_\textrm{ASE}(\boldsymbol{d}) \big| - \frac{2}{m} \sum_{i=1}^{l} \sum_{j=1}^{m} d_i \log \lambda_j \right\} = A_1 + A_2,
\end{equation}
with
\[
\begin{aligned}
	A_1 =&~ \sum_{i=1}^{l} \eta_i \tr\left[ \frac{\widehat{G}_{\mathrm{ASE}}(\boldsymbol{d}^0)^{-1}}{\sqrt{m}} \sum_{j=1}^{m} \log \lambda_j \re\left[ \Psi_j(\boldsymbol{d}^0)^{-1} \left( \mathbf{J}_{:,i}(\lambda_j) + \mathbf{J}_{i,:}(\lambda_j) \right) \overline{\Psi_j(\boldsymbol{d}^0)^{-1}} \right]  \right] \\
	&- \sum_{j=1}^{m} \sum_{i=1}^{l} \frac{2\eta_i \log \lambda_j}{\sqrt{m}}
\end{aligned}
\]
and
\[
A_2 = \sum_{i=1}^{l} \eta_i \tr\left[ \frac{\widehat{G}_{\mathrm{ASE}}(\boldsymbol{d}^0)^{-1}}{\sqrt{m}} \sum_{j=1}^{m} \frac{\lambda_j-\pi}{2} \im\left[ \Psi_j(\boldsymbol{d}^0)^{-1} \left( -\mathbf{J}_{:,i}(\lambda_j) + \mathbf{J}_{i,:}(\lambda_j) \right) \overline{\Psi_j(\boldsymbol{d}^0)^{-1}} \right] \right] .
\]
We proceed to find an approximation of $A_1$ and $A_2$. Provided that
\[
\mu_j = \log \lambda_j - \frac{1}{m} \sum_{j=1}^{m} \log \lambda_j = \log j - \frac{1}{m} \sum_{j=1}^{m} \log j = \mathcal{O}(\log m) ,
\] 
then we have
\[
\resizebox{1\hsize}{!}{$
	\begin{aligned}
		A_1 &= \sum_{i=1}^{l} \eta_i \left\{ \tr\left[ \widehat{G}_{\mathrm{ASE}}(\boldsymbol{d}^0)^{-1} \frac{1}{\sqrt{m}} \sum_{j=1}^{m} \log \lambda_j \re\left[ \Psi_j(\boldsymbol{d}^0)^{-1} \left( \mathbf{J}_{:,i}(\lambda_j) + \mathbf{J}_{i,:}(\lambda_j) \right) \overline{\Psi_j(\boldsymbol{d}^0)^{-1}} \right]  \right] - \sum_{j=1}^{m} \frac{2 \log \lambda_j}{\sqrt{m}} \right\} \\
		&= \sum_{i=1}^{l} \eta_i \tr\left[ \widehat{G}_{\mathrm{ASE}}(\boldsymbol{d}^0)^{-1} \frac{2}{\sqrt{m}} \sum_{j=1}^{m} \mu_j \re\left[ \Psi_j(\boldsymbol{d}^0)^{-1} \mathbf{J}_{:,i}(\lambda_j) \overline{\Psi_j(\boldsymbol{d}^0)^{-1}} \right] \right]\\
		&= \sum_{i=1}^{l} \eta_i \left\{ \frac{2}{\sqrt{m}} \sum_{j=1}^{m} \mu_j \sum_{k=1}^{l} \left( g^{-1}_{i,k} + \op(1) \right) \re\left[ \lambda_j^{d_0} \e^{\ii (\lambda_j - \pi) d_0/2} \phi_{k,i} \lambda_j^{d_0} \e^{\ii (\pi - \lambda_j) d_0/2} \right] \right\} \\
		&= \sum_{i=1}^{l} \eta_i \left\{ \frac{2}{\sqrt{m}} \sum_{j=1}^{m} \left[ \mu_j \sum_{k=1}^{l} \left( g^{-1}_{i,k} + \op(1) \right) \re\left[ \lambda_j^{d_0} \e^{\ii (\lambda_j - \pi) d_0/2} \phi_{k,i}' \lambda_j^{d_0} \e^{\ii (\pi - \lambda_j) d_0/2} \right]  + \op(\sqrt{m}) \right] \right\} \\
		&= \frac{2}{\sqrt{m}} \sum_{i=1}^{l} \eta_i \sum_{j=1}^{m} \mu_j \sum_{k=1}^{l} \left\{ g_{i,k}^{-1} \re\left[ \lambda_j^{d_0} \e^{\ii (\lambda_j - \pi) d_0/2} \phi_{k,i}' \lambda_j^{d_0} \e^{\ii (\pi - \lambda_j) d_0/2} \right] - 1 \right\} + \op(1) ,\\
	\end{aligned} $}
\]
where $g_{i,k}$, $\phi_{i,k}$, and $\phi_{i,k}'$ denote the $(i,k)$-element of matrices $\widehat{G}_{\mathrm{ASE}}(\boldsymbol{d}^0)$, $\mathbf{J}_T(\lambda_j)$, and $\mathbf{I}_T(\lambda_j)$, respectively. The second equality establishes by multiplying $\sum_{j=1}^{m} \log \lambda_j / \sqrt{m}$ by the multiplier $\widehat{G}_{\mathrm{ASE}}(\boldsymbol{d}^0)^{-1} \widehat{G}_{\mathrm{ASE}}(\boldsymbol{d}^0)$, the third and fourth equalities follow from Lemma~\ref{tec:an}, and the last equality holds according to
\[
\begin{aligned}
	&\frac{1}{\sqrt{m}} \sum_{j=1}^{m} \mu_j \lambda_j^{d_0} \e^{\ii (\lambda_j - \pi) d_0/2} \phi_{i,k} \lambda_j^{d_0} \e^{\ii (\pi - \lambda_j) d_0/2} \\
	&\quad= \frac{1}{\sqrt{m}} \sum_{j=1}^{m} \mu_j \lambda_j^{d_0} \e^{\ii (\lambda_j - \pi) d_0/2} \left( \phi_{i,k}' + \op\left(\frac{\sqrt{m}}{\log m}\right) \right) \lambda_j^{d_0} \e^{\ii (\pi - \lambda_j) d_0/2}
	\quad\left( \text{From Lemma~\ref{tec:an}} \right) \\
	&\quad= \frac{1}{\sqrt{m}} \sum_{j=1}^{m} \mu_j \lambda_j^{d_0} \e^{\ii (\lambda_j - \pi) d_0/2} \phi_{i,k}' \lambda_j^{d_0} \e^{\ii (\pi - \lambda_j) d_0/2} + \frac{1}{\sqrt{m}} \op(\sqrt{m}) \\
	&\quad= \frac{1}{\sqrt{m}} \sum_{j=1}^{m} \mu_j \left[ \lambda_j^{d_0} \e^{\ii (\lambda_j - \pi) d_0/2} \phi_{i,k}' \lambda_j^{d_0} \e^{\ii (\pi - \lambda_j) d_0/2} - G^0_{i,k} \right] + \op(1) ,\\
\end{aligned}
\]
provided $\sum_{j=1}^{m} \mu_j = 0$ and Lemma~\ref{tec:an}. Hence, we have proved that
\begin{equation} \label{eq:A1}
	A_1 = \frac{2}{\sqrt{m}} \sum_{i=1}^{l} \eta_i \sum_{j=1}^{m} \mu_j \sum_{k=1}^{l} \left\{ g_{i,k}^{-1} \re\left[ \lambda_j^{d_0} \e^{\ii (\lambda_j - \pi) d_0/2} \phi_{i,k}' \lambda_j^{d_0} \e^{\ii (\pi - \lambda_j) d_0/2} \right] - 1 \right\} + \op(1) .
\end{equation}
Similarly, we can obtain
\begin{equation} \label{eq:A2}
	A_2 = - \frac{\pi}{\sqrt{m}} \sum_{i=1}^{l} \eta_i \sum_{j=1}^{m} \mu_j \sum_{k=1}^{l} \im\left[ \lambda_j^{-d_i} \e^{\ii (\lambda_j - \pi) d_i/2} \phi_{i,k}'  \lambda_j^{-d_i} \e^{\ii (\pi - \lambda_j) d_i/2} \right] + \op(1) .
\end{equation}
Therefore, Eq.~\eqref{eq:an_1} holds as the $\op(1)$ in Eqs.~\eqref{eq:A1} and~\eqref{eq:A2} converges in probability to zero as $T \to \infty$.

\noindent\textbf{Hessian approximation.} Recall several notations in this proof
\[
\Omega_{\delta} = \{ \boldsymbol{d} \mid \| \boldsymbol{d} - \boldsymbol{d}^0 \|_{\infty} > \delta \} \quad\text{and}\quad \Omega_{\delta,T} = \left\{ \boldsymbol{d} \mid (\log T)^4 \| \boldsymbol{d} - \boldsymbol{d}^0 \|_{\infty} < \delta \right\}, \quad\text{for}\quad \delta>0 .
\]
Following the proof of ``\textbf{About $\boldsymbol{P_1}$}", for $\boldsymbol{d} \in \Omega_{\beta}$ and $\Delta\boldsymbol{d} \in \Omega_1$, one has
\[
\left\{\begin{aligned}
	& \inf_{\Omega_1 \setminus \Omega_{\delta,T}} S_r(\boldsymbol{d}) \geq \delta^2(\log T)^8 >0 ,\\
	& |A(\boldsymbol{d}) - H(\boldsymbol{d})| = \op \left( m^{\beta}T^{-\beta} + m^{-2\delta} \log m + mT^{-1} \right) ,\\
	& \log A(\boldsymbol{d}) - \log B(\boldsymbol{d}) \geq \log \left( H(\boldsymbol{d}) + \op\left( (\log T)^8 \right) \right) - \log H(\boldsymbol{d}) = \op\left( (\log T)^{-8} \right) ,\\
	& \log A(\boldsymbol{d}^0) - \log B(\boldsymbol{d}^0) = \log \left( H(\boldsymbol{d}^0) + \op\left( (\log T)^8 \right) \right) - \log H(\boldsymbol{d}^0) = \op\left( (\log T)^{-8} \right) ,
\end{aligned} \right.
\]
where
\[
\left\{\begin{aligned}
	A(\boldsymbol{d}) &= \left( \frac{2 \pi m}{T} \right)^{-2 \sum_i d'_i } \left| \widehat{G}_{\mathrm{ASE}}(\boldsymbol{d}) \right| ,\\
	B(\boldsymbol{d}) &= \prod_{i=1}^{l} \frac{1}{2d'_i + 1} \left| \widehat{G}^0 \right| ,\\
	S_r(\boldsymbol{d}) &= - 2 \sum_{i=1}^{l} d'_i \left( \frac{1}{m} \sum_{j=1}^{m} \log j - \log m \right) - \sum_{i=1}^{l} \log(2 d'_i + 1) ,\\
	H(\boldsymbol{d}) &= \Big| \re[\mathbf{U}(\Delta\boldsymbol{d})] \odot G^0 \odot \mathbf{V}(\Delta\boldsymbol{d}) \Big| .
\end{aligned} \right.
\]
Thus, we have
\[
\mathbb{P} \left( \inf _{\Omega_1 \setminus \Omega_{\delta,T}} S(\boldsymbol{d}) \leq 0 \right) \to 0 \quad\text{and}\quad  \mathbb{P}\left( \Delta\boldsymbol{d} \in \Omega_{\delta,T} \right) \to 1 \quad\text{as}\quad T \to \infty ,
\]
where
\[
\begin{aligned}
	S(\boldsymbol{d}) &= \log \left| \widehat{G}_{\mathrm{ASE}}(\boldsymbol{d}) \right| - \log \left| \widehat{G}_{\mathrm{ASE}}( \boldsymbol{d}^0 ) \right| - \frac{2}{m} \sum_{i=1}^{l} \sum_{j=1}^{m} d'_i \log \lambda_j \\
	&= \left[ \log A(\boldsymbol{d}) - \log B(\boldsymbol{d}) \right] - \left[ \log A(\boldsymbol{d}^0) - \log B(\boldsymbol{d}^0) \right] + S_r(\boldsymbol{d}) .
\end{aligned}
\]
For $p,q \in [l]$, it is observed that
\begin{equation} \label{eq:an_hessian}
	\frac{\partial^2 }{\partial d_p \partial d_q } \left( \log \big| \widehat{G}_\textrm{ASE} \big| - \frac{2}{m} \sum_{i=1}^{l} \sum_{j=1}^{m} d_i \log \lambda_j \right) = \tr \left[ -\widehat{G}_{\mathrm{ASE}}^{-1} \frac{\partial \widehat{G}_{\mathrm{ASE}}}{\partial d_p} \widehat{G}_{\mathrm{ASE}}^{-1} \frac{\partial \widehat{G}_{\mathrm{ASE}}}{\partial d_q} + \widehat{G}_{\mathrm{ASE}}^{-1} \frac{\partial^2 \widehat{G}_\textrm{ASE} }{\partial d_p \partial d_q } \right]
\end{equation}
with
\begin{equation} \label{eq:an_first}
	\begin{aligned}
		& \frac{\partial \widehat{G}_{\mathrm{ASE}}}{\partial d_p} = \frac{1}{m} \sum_{j=1}^{m} \re\left[ \left(\log\lambda_j + \ii\frac{\pi-\lambda_j}{2}\right) \Psi_j(\boldsymbol{d}^0)^{-1} \mathbf{J}_{:,p}(\lambda_j) \overline{\Psi_j(\boldsymbol{d}^0)^{-1}} \right] \\
		&\quad\quad\quad\quad~~ + \frac{1}{m} \sum_{j=1}^{m} \re\left[ \left(\log\lambda_j + \ii\frac{\lambda_j-\pi}{2}\right) \Psi_j(\boldsymbol{d}^0)^{-1} \mathbf{J}_{p,:}(\lambda_j) \overline{\Psi_j(\boldsymbol{d}^0)^{-1}} \right]  \\
	\end{aligned}
\end{equation}
and
\begin{equation} \label{eq:an_second}
	\begin{aligned}
		& \frac{\partial^2 \widehat{G}_\textrm{ASE}}{\partial d_p \partial d_q} = \frac{1}{m} \sum_{j=1}^{m} \re\left[ \left(\log\lambda_j + \ii\frac{\pi-\lambda_j}{2}\right)^2 \Psi_j(\boldsymbol{d}^0)^{-1} \widehat{\mathbf{J}_{p,q}}(\lambda_j) \overline{\Psi_j(\boldsymbol{d}^0)^{-1}} \right] \\
		&\quad\quad\quad\quad~~ + \frac{1}{m} \sum_{j=1}^{m} \re\left[ \left(\log\lambda_j + \ii\frac{\lambda_j-\pi}{2} \right)^2 \Psi_j(\boldsymbol{d}^0)^{-1} \widehat{\mathbf{J}_{p,q}}(\lambda_j) \overline{\Psi_j(\boldsymbol{d}^0)^{-1}} \right] \\
		&\quad\quad\quad\quad~~ + \frac{1}{m} \sum_{j=1}^{m} \re\left[ \left|\log\lambda_j+\ii\frac{\pi-\lambda_j}{2}\right|^2 \Psi_j(\boldsymbol{d}^0)^{-1} \mathbf{J}_{p,q}(\lambda_j) \overline{\Psi_j(\boldsymbol{d}^0)^{-1}} \right] \\
		&\quad\quad\quad\quad~~ + \frac{1}{m} \sum_{j=1}^{m} \re\left[ \left|\log\lambda_j+\ii\frac{\lambda_j-\pi}{2}\right|^2 \Psi_j(\boldsymbol{d}^0)^{-1} \mathbf{J}_{q,p}(\lambda_j) \overline{\Psi_j(\boldsymbol{d}^0)^{-1}} \right] ,
	\end{aligned}
\end{equation}
where $\mathbf{J}_{p,q}(\lambda_j)$ denotes a $l \times l$ matrix whose $(p,q)$-th element is the same as that of $\mathbf{J}_T(\lambda_j)$ while all other elements are zero, $\widehat{\mathbf{J}_{p,q}}(\lambda_j) = \mathbf{J}_{:,p=q}(\lambda_j)$ and $\widehat{\mathbf{J}_{q,p}}(\lambda_j) = \mathbf{J}_{p=q,:}(\lambda_j)$ for $p=q$, otherwise, $\widehat{\mathbf{J}_{p,q}}(\lambda_j) = \mathbf{0}$, and here we omit the $\boldsymbol{d}$ in $\widehat{G}_{\mathrm{ASE}}(\boldsymbol{d})$ for brevity,

Altering to the proof line of Eq.~\eqref{eq:use1}, we have
\begin{equation} \label{eq:an_use1}
	\begin{aligned}
		&\re\left[ \left(\log\lambda_j + \ii\frac{\pi-\lambda_j}{2}\right) \Psi_j(\boldsymbol{d}^0)^{-1} \mathbf{J}_{:,p}(\lambda_j) \overline{\Psi_j(\boldsymbol{d}^0)^{-1}} \right] \\
		&\quad= \bigg( \log\lambda_j + \ii\frac{\pi-\lambda_j}{2} \bigg) \bigg( \re\left[ \Psi_j(\boldsymbol{d}^0)^{-1} \mathbf{J}_{:,p}(\lambda_j) \overline{\Psi_j(\boldsymbol{d}^0)^{-1}} \right] + \ii \im\left[ \Psi_j(\boldsymbol{d}^0)^{-1} \mathbf{J}_{:,p}(\lambda_j) \overline{\Psi_j(\boldsymbol{d}^0)^{-1}} \right] \bigg)    \\
		&\quad= \log \lambda_j \re\left[ \Psi_{j}(\boldsymbol{d})^{-1} \mathbf{J}_{:,p}(\lambda_j) \overline{\Psi_{j}(\boldsymbol{d})}^{-1} \right] + \frac{\pi - \lambda_j}{2} \im \left[ \Psi_{j}(\boldsymbol{d})^{-1} \mathbf{J}_{:,p}(\lambda_j) \overline{\Psi_{j}(\boldsymbol{d})}^{-1} \right] .
	\end{aligned}
\end{equation}
Given $\alpha \in (0,2]$ in Assumption~\ref{ass:B1}, we have
\begin{equation} \label{eq:an_use2}
	\begin{aligned}
		& \frac{1}{m} \sum_{j=1}^{m} \lambda_j \Psi_{j}(\boldsymbol{d})^{-1} \mathbf{J}_T(\lambda_j) \overline{\Psi_{j}(\boldsymbol{d})}^{-1} \\
		&\quad\leq \frac{1}{m} \sum_{j=1}^{m-1} \left| \lambda_j - \lambda_{j+1} \right| \left\| \sum_{k=1}^{j} \Psi_{j}(\boldsymbol{d})^{-1} \mathbf{J}_T(\lambda_j) \overline{\Psi_{j}(\boldsymbol{d})}^{-1} \right\|_{\infty}
		+ \frac{\lambda_m}{m} \left\| \sum_{k=1}^{j} \Psi_{j}(\boldsymbol{d})^{-1} \mathbf{J}_T(\lambda_j) \overline{\Psi_{j}(\boldsymbol{d})}^{-1} \right\|_{\infty} \\
		&\quad\leq \frac{m-1}{mT} \left[ \op\left( \sqrt{m-1} \log(m-1) + \frac{(m-1)^{\alpha+1}}{T^{\alpha}} \right) + \mathcal{O} \left( \frac{1}{m} \right) \right] \\
		&\quad\quad+ \mathcal{O}\left( \frac{1}{n} \right) \left[ \op \left( \sqrt{m} \log m + \frac{m^{\alpha+1}}{T^{\alpha}} \right) + \mathcal{O} \left( \frac{1}{m} \right) \right] \\
		&\quad= \op\left( \frac{\sqrt{m} \log m}{T} + \frac{m^{\alpha}}{T^{\alpha}} \right) + \mathcal{O} \left( \frac{1}{mT} \right) \\
		&\quad= \op\left( \frac{\sqrt{m} }{ \sqrt{T}} \frac{\log m}{T^{1/4}} \frac{1}{T^{1/4}} \right) + \op\left( \frac{m^{\alpha}}{T^{\alpha}} \right) + \mathcal{O} \left( \frac{1}{mT} \right)
		\quad\left(\text{From Lemma~\ref{tec:our1}},~\text{as}~ T\to\infty \right) \\
		&\quad= \op \left( \frac{1}{\log T} \right) .
	\end{aligned}
\end{equation}
Invoking Eq.~\eqref{eq:an_use2} into Eq.~\eqref{eq:an_use1}, we can bound the first term of Eq.~\eqref{eq:an_first}. Similarly, we can bound its second term, that is, $m^{-1} \sum_{j=1}^{m} \re\left[ \left(\lambda_j + \ii\frac{\pi-\lambda_j}{2}\right) \Psi_j(\boldsymbol{d}^0)^{-1} \mathbf{J}_{p,:}(\lambda_j) \overline{\Psi_j(\boldsymbol{d}^0)^{-1}} \right]$. Thus, we have
\begin{equation} \label{eq:an_c1}
	\begin{aligned}
		\frac{\partial \widehat{G}_{\mathrm{ASE}}}{\partial d_p} =&~ \frac{1}{m} \sum_{j=1}^{m} \left\{ \re\left[ \log\lambda_j \Psi_j(\boldsymbol{d}^0)^{-1} \mathbf{J}_{:,p}(\lambda_j) \overline{\Psi_j(\boldsymbol{d}^0)^{-1}} \right] + \re\left[ \log\lambda_j \Psi_j(\boldsymbol{d}^0)^{-1} \mathbf{J}_{p,:}(\lambda_j) \overline{\Psi_j(\boldsymbol{d}^0)^{-1}} \right] \right\} \\
		&+ \frac{\pi}{2m} \sum_{j=1}^{m} \left\{ \im\left[ \Psi_j(\boldsymbol{d}^0)^{-1} \mathbf{J}_{:,p}(\lambda_j) \overline{\Psi_j(\boldsymbol{d}^0)^{-1}} \right] - \im\left[ \Psi_j(\boldsymbol{d}^0)^{-1} \mathbf{J}_{p,:}(\lambda_j) \overline{\Psi_j(\boldsymbol{d}^0)^{-1}} \right] \right\} \\
		&+ \op\left( \frac{1}{\log T} \right) .
	\end{aligned}
\end{equation}
On the other hand, given $\alpha \in (0,2]$ in Assumption~\ref{ass:B1}, we have
\begin{equation} \label{eq:an_use3}
	\begin{aligned}
		&\left\| \frac{1}{m} \sum_{j=1}^{m} \lambda_j \log \lambda_j \Psi_{j}(\boldsymbol{d})^{-1} \mathbf{J}_T(\lambda_j) \overline{\Psi_{j}(\boldsymbol{d})}^{-1} \right\|_{\infty} \\
		&\quad\quad\leq \frac{1}{m} \sum_{j=1}^{m-1} \Big| \lambda_j \log \lambda_j - \lambda_{j+1} \log \lambda_{j+1} \Big| \left\| \sum_{k=1}^{j} \Psi_{j}(\boldsymbol{d})^{-1} \mathbf{J}_T(\lambda_j)  \overline{\Psi_{j}(\boldsymbol{d})}^{-1} \right\|_{\infty} \\
		&\quad\quad\quad+ \frac{\lambda_m \log \lambda_m}{m} \left\| \sum_{k=1}^{j} \Psi_{j}(\boldsymbol{d})^{-1} \mathbf{J}_T(\lambda_j) \overline{\Psi_{j}(\boldsymbol{d})}^{-1} \right\|_{\infty} \\
		&\quad\quad\leq \op \left( \frac{\sqrt{m} \log(m-1)}{m} + \frac{(m-1)^{\alpha+1}}{mT^{\alpha}} \right) o(1) + \op(1) \\
		&\quad\quad\leq \op(1) .
	\end{aligned}
\end{equation}
Invoking Eq.~\eqref{eq:an_use3} into Eq.~\eqref{eq:an_second}, we have
\begin{equation} \label{eq:an_c2}
	\begin{aligned}
		\frac{\partial^2 \widehat{G}_\textrm{ASE}}{\partial d_p \partial d_q}
		&= \frac{\pi^2}{4m} \sum_{j=1}^{m} \left\{ \re\left[ \Psi_j(\boldsymbol{d}^0)^{-1} \widehat{\mathbf{J}_{p,q}}(\lambda_j) \overline{\Psi_j(\boldsymbol{d}^0)^{-1}} \right]
		+ \re\left[ \Psi_j(\boldsymbol{d}^0)^{-1} \widehat{\mathbf{J}_{q,p}}(\lambda_j) \overline{\Psi_j(\boldsymbol{d}^0)^{-1}} \right] \right\} \\
		&\quad+ \frac{\pi^2}{4m} \sum_{j=1}^{m} \left\{ \re\left[ \Psi_j(\boldsymbol{d}^0)^{-1} \mathbf{J}_{p,q}(\lambda_j) \overline{\Psi_j(\boldsymbol{d}^0)^{-1}} \right]
		+ \re\left[ \Psi_j(\boldsymbol{d}^0)^{-1} \mathbf{J}_{q,p}(\lambda_j) \overline{\Psi_j(\boldsymbol{d}^0)^{-1}} \right] \right\} \\
		&\quad+ \frac{\pi}{m} \sum_{j=1}^{m} \log \lambda_j \left\{ \im\left[ \Psi_j(\boldsymbol{d}^0)^{-1} \widehat{\mathbf{J}_{p,q}}(\lambda_j) \overline{\Psi_j(\boldsymbol{d}^0)^{-1}} \right]
		+ \im\left[ \Psi_j(\boldsymbol{d}^0)^{-1} \widehat{\mathbf{J}_{q,p}}(\lambda_j) \overline{\Psi_j(\boldsymbol{d}^0)^{-1}} \right] \right\} \\
		&\quad+ \frac{1}{m} \sum_{j=1}^{m} (\log \lambda_j)^2 \left\{ \re\left[ \Psi_j(\boldsymbol{d}^0)^{-1} \widehat{\mathbf{J}_{p,q}}(\lambda_j) \overline{\Psi_j(\boldsymbol{d}^0)^{-1}} \right]
		+ \re\left[ \Psi_j(\boldsymbol{d}^0)^{-1} \widehat{\mathbf{J}_{q,p}}(\lambda_j) \overline{\Psi_j(\boldsymbol{d}^0)^{-1}} \right] \right\} \\
		&\quad+ \frac{1}{m} \sum_{j=1}^{m} (\log \lambda_j)^2 \left\{ \re\left[ \Psi_j(\boldsymbol{d}^0)^{-1} \mathbf{J}_{p,q}(\lambda_j) \overline{\Psi_j(\boldsymbol{d}^0)^{-1}} \right]
		+ \re\left[ \Psi_j(\boldsymbol{d}^0)^{-1} \mathbf{J}_{q,p}(\lambda_j) \overline{\Psi_j(\boldsymbol{d}^0)^{-1}} \right] \right\} \\
		&\quad+ \op(1) .
	\end{aligned}
\end{equation}
Notice that both ${\partial \widehat{G}_{\mathrm{ASE}}} / {\partial d_p}$ and ${\partial^{2} \widehat{G}_\textrm{ASE} } / {\partial d_p \partial d_q }$ have the reminders $\op \left( (\log T)^{-1} \right)$ and $\op(1)$, respectively. Invoking Eqs.~\eqref{eq:an_c1} and~\eqref{eq:an_c2} into Eq.~\eqref{eq:an_hessian}, it is sufficient to show the limit approximation about two terms $\re\left[ \Psi_j(\boldsymbol{d}^0)^{-1} \mathbf{J}_T(\lambda_j) \overline{\Psi_j(\boldsymbol{d}^0)^{-1}} \right]$ and $\im\left[ \Psi_j(\boldsymbol{d}^0)^{-1} \mathbf{J}_T(\lambda_j) \overline{\Psi_j(\boldsymbol{d}^0)^{-1}} \right]$. 

For $k=0,1,2$, we define
\[
R_k(\boldsymbol{d}) = \frac{1}{m} \sum_{j=1}^{m} (\log \lambda_j)^k \re\left[ \Psi_j(\Delta\boldsymbol{d})^{-1} G^0 \overline{\Psi_j(\Delta\boldsymbol{d})^{-1}} \right] .
\]
It's observed that for $\Delta\boldsymbol{d} \in \Omega_{\delta,T}$,
\begin{equation}  \label{eq:final1}
	\begin{aligned}
		& \sup_{\Omega_{\delta,T}} \left\|\frac{1}{m} \sum_{j=1}^{m} \log \left(\lambda_j\right)^k \Psi_j(\boldsymbol{d})^{-1} \mathbf{J}_T \left(\lambda_j\right) \overline{\Psi_j(\boldsymbol{d})^{-1}} - R_k(\Delta\boldsymbol{d}) \right\|_{\infty} \\
		&\quad= \sup_{\Omega_{\delta,T}} \left\|\frac{1}{m} \sum_{j=1}^{m} \log \left(\lambda_j\right)^k \Psi_j(\Delta\boldsymbol{d})^{-1} \left( \Psi_j(\boldsymbol{d}^0)^{-1} \mathbf{J}_T\left(\lambda_j\right) \overline{\Psi_j(\boldsymbol{d}^0)^{-1}} - G^0 \right) \overline{\Psi_j(\Delta\boldsymbol{d})^{-1}} \right\|_{\infty} \\
		&\quad= \op\left( \frac{(\log T)^k}{m} \sum_{j=1}^{m} \left(j^{\beta} T^{-\beta} + j^{-1/2} \log j\right)\right) 
		\quad\left(\text{From Lemma~\ref{tec:our2}},~\text{as}~ T\to\infty \right)  \\
		&\quad= \op\left( (\log T)^{k-2} \right) 
	\end{aligned}
\end{equation}
and
\[
\begin{aligned}
	& \frac{1}{m} \sum_{j=1}^{m} \log \left(\lambda_j\right)^k \left[ \e^{\ii (\lambda_j - \pi)(d'_p - d'_q)/2} \lambda_j^{d'_p+d'_q} - 1 \right] G^0_{p,q} \\
	&\quad\leq \frac{1}{m} \sum_{j=1}^{m} \log \left(\lambda_j\right)^k \sup_{\Omega_{\delta,T}} \left| \e^{\ii (\lambda_j - \pi)(d'_p - d'_q)/2} \lambda_j^{d'_p+d'_q} - 1 \right| G^0_{p,q} \\
	&\quad\leq \frac{1}{m} \sum_{j=1}^{m} \log \left(\lambda_j\right)^k \sup_{\Omega_{\delta,T}} C\left(|d'_p|+|d'_q|\right) G^0_{p,q} 
	\quad\left(\text{From Lemma~\ref{tec:our2}},~\text{for}~ C > 1 \right) \\
	&\quad= \frac{1}{m} \sum_{j=1}^{m} \log \left(\lambda_j\right)^k \mathcal{O} \left( (\log T)^{-3} \right) G^0_{p,q} .
\end{aligned}
\]
Since the formula above corresponds to the $(p,q)$-th element, we have
\begin{equation}  \label{eq:final2}
	\sup_{\Omega_{\delta,T}} \left\| \frac{1}{m} \sum_{j=1}^{m} \log \left(\lambda_j\right)^k G^0 - R_k(\Delta\boldsymbol{d}) \right\|_{\infty} = o\left( (\log T)^{k-2} \right) .
\end{equation}
Based on Eqs.~\eqref{eq:final1} and \eqref{eq:final2}, for $\Delta\boldsymbol{d} \in \Omega_{\delta,T}$, we have
\begin{equation} \label{eq:an_c3}
	\frac{1}{m} \sum_{j=1}^{m} (\log \lambda_j)^k \re\left[ \Psi_j(\boldsymbol{d}^0)^{-1} \mathbf{J}_T(\lambda_j) \overline{\Psi_j(\boldsymbol{d}^0)^{-1}} \right] = G^0 \frac{1}{m} \sum_{j=1}^{m} (\log \lambda_j)^k + \op\left( (\log T)^{k-2} \right)
\end{equation}
and
\begin{equation} \label{eq:an_c4}
	\frac{1}{m} \sum_{j=1}^{m} (\log \lambda_j)^k \im\left[ \Psi_j(\boldsymbol{d}^0)^{-1} \mathbf{J}_T(\lambda_j) \overline{\Psi_j(\boldsymbol{d}^0)^{-1}} \right] = \op\left( (\log T)^{k-2} \right) .
\end{equation}
Finally, invoking Eqs.~\eqref{eq:an_c1}, \eqref{eq:an_c2}, \eqref{eq:an_c3}, and \eqref{eq:an_c4} into Eq.~\eqref{eq:an_hessian}, we obtain
\[
\frac{\partial^2 }{\partial d_p \partial d_q } \left( \log \big| \widehat{G}_\textrm{ASE} \big| - \frac{2}{m} \sum_{i=1}^{l} \sum_{j=1}^{m} d_i \log \lambda_j \right) = \tr\left[ R_1(G^0) + R_2(G^0) \right] + \op(1) ,
\]
where for $p,q \in [l]$,
\[
\left\{\begin{aligned}
	R_1(G^0) &= \left(G^0\right)^{-1} \left( \widehat{\mathbf{G}_{p,q}} + \widehat{\mathbf{G}_{q,p}} + \mathbf{G}_{p,q} + \mathbf{G}_{q,p} \right) ,\\
	R_2(G^0) &= \frac{\pi^2}{4} \left(G^0\right)^{-1} \left( \mathbf{G}_{p,q} + \mathbf{G}_{q,p} - \widehat{\mathbf{G}_{p,q}} - \widehat{\mathbf{G}_{q,p}} \right) ,\\
\end{aligned}\right.
\]
in which $\mathbf{G}_{p,q}(\lambda_j)$ denotes a $l \times l$ matrix whose $(p,q)$-th element is $g_{p,q}$ while all other elements are zero, and if $p=q$, the $p$-th (or $q$-th) column and row elements of matrices $\widehat{\mathbf{G}_{p,q}}$ and $\widehat{\mathbf{G}_{q,p}}$ are filled with $( g_{1,p=q}, \dots, g_{l,p=q} )^{\top}$ and $( g_{p=q,1}, \dots, g_{p=q,l} )$, respectively, whereas all other elements are zero; if $p \neq q$, $\widehat{\mathbf{G}_{p,q}} = \mathbf{0}$. This completes the proof.  $\hfill\square$

\section{GSE and TSE}  \label{app:GSE}
Recall the aforementioned consensus of the multivariate fractionally integrated process, in which there exists a symmetric and positive-definite matrix $G \in \mathbb{R}^{l \times l}$ such that
\[
f_X(\lambda) = \Lambda(\lambda) ~f_{\epsilon}(\lambda)~ \overline{\Lambda(\lambda)} \quad\text{and}\quad f_{\epsilon}(\lambda) \sim G , \quad\text{with}\quad \Lambda(\lambda) = \diag_{i \in\{1, \cdots, l\}}\{ (1-\mathfrak{B})^{-d_i} \} .
\]
In general, the estimation value of $\boldsymbol{d}$ can be empirically calculated by maximizing the following Gaussian log-likelihood function localized to the origin
\begin{equation} \label{eq:LL}
	LL_m(G, \boldsymbol{d}) = 
	\frac{1}{m} \sum_{j=1}^{m} \bigg\{ \log \Big| \Lambda(\lambda_j) ~G~ \overline{\Lambda(\lambda_j)} \Big| + \tr\left[ \left(\Lambda(\lambda_j) ~G~ \overline{\Lambda(\lambda_j)} \right)^{-1} f_X\left(\lambda_j\right) \right] \bigg\} ,
\end{equation}
where $m = |\{\lambda_j\}|_{\#}$ denotes the number of empirical frequencies $\{\lambda_j\}$. It's observed that solving Eq.~\eqref{eq:LL} relies on the calculations of $\Lambda(\lambda_j)$ and $f_X(\lambda_j)$ for $j\in[m]$. To solve these issues, Lobato~\citep{lobato1999:two} decomposes Eq.~\eqref{eq:starting} into a univariate formation, i.e., $(1-\mathfrak{B})^{d_i} \mathbf{X}_{i t} = \epsilon_{i t}$ for $i \in[l]$ and employs the periodogram estimation $\mathbf{I}_T$ described by Eq.~\eqref{eq:I} for estimating $f_X$. These manners leads to a two-step approach, including a first-step univariate estimation of $d_i$ and a Newton-type step. Formally, we have
\begin{equation} \label{eq:two}
	\mathbf{I}_T (\lambda_j) \approx f_X(\lambda_j) \sim \Phi_{j}(\boldsymbol{d}) ~\widehat{G}~ \overline{\Phi_{j}(\boldsymbol{d})}^{\top} \quad\text{with}\quad \Phi_{j}(\boldsymbol{d}) = \diag_{i \in\{1, \cdots, l\}} \left\{ \lambda_j^{-d_i} \right\} \in \mathbb{R}^{l \times l} .
\end{equation}
The consistency and asymptotic normality of the two-step approach~\citep{lobato1999:two} hold beyond the component Gaussian assumption that each component sequence $\{\epsilon_{it}\}_{t=0}^{\infty}$ belongs to a Gaussian process for $i\in[l]$.

Alternatively, Shimotsu~\citep{shimotsu2007:GSE} develops the \textit{Gaussian Semi-parametric Estimator} (GSE), which considers a more general local form as follows
\begin{equation} \label{eq:GSE_f}
	f_X(\lambda_j) \sim \Psi_{j}(\boldsymbol{d}) ~\widehat{G}~ \overline{\Psi_{j}(\boldsymbol{d})}^{\top} \quad\text{with}\quad \Psi_{j}(\boldsymbol{d}) = \diag_{i \in\{1, \cdots, l\}} \left\{ \lambda_j^{-d_i} \e^{\ii\left(\pi-\lambda_j\right) d_i / 2}  \right\} \in \mathbb{C}^{l \times l}.
\end{equation}
In comparison with Eq.~\eqref{eq:two}, the GSE has a more precise expansion of the operator $\overline{\Lambda(\lambda)}$ according to
\[
\left(1-e^{\ii \lambda}\right)^{d_i} = \lambda^{d_i} \e^{\ii(\lambda-\pi) d_i / 2} \left(1 + \mathcal{O}\left(\lambda^{2}\right)\right), \quad i\in[l] .	
\]
Since $\arg(1-\e^{\ii\lambda}) = (\lambda - \pi) /2$ for $\lambda \in [0,\pi]$, the multiplier $\e^{\ii(\lambda-\pi) d_i / 2}$ intrinsically indicates the phase of operator $\overline{\Lambda(\lambda)}$ for $i\in[l]$. This refinement of the GSE not only allows the rotation operations between component time series, relaxing the component Gaussian assumption to Gaussian joint priors for $\boldsymbol{\epsilon}_t$, but also provides a smaller limiting variance than that of Eq.~\eqref{eq:two}. Invoking Eqs.~\eqref{eq:I} and~\eqref{eq:GSE_f} into Eq.~\eqref{eq:LL}, the GSE estimator $\widehat{\boldsymbol{d}}_{\textrm{GSE}}$ can be obtained by solving the optimization
\begin{equation} \label{eq:GSE}
	\widehat{\boldsymbol{d}}_{\textrm{GSE}} = \arg\min_{\boldsymbol{d}} \left\{ \log \bigg| \frac{1}{m} \sum_{j=1}^{m} \re\left[\Psi_{j}(\boldsymbol{d})^{-1} \mathbf{I}_T(\lambda_j) \overline{\Psi_{j}(\boldsymbol{d})}^{-1}\right] \bigg| + \frac{1}{m} \sum_{i=1}^{l} \log \bigg| \Psi_{j}(\boldsymbol{d})^{-1} \overline{\Psi_{j}(\boldsymbol{d})}^{-1} \bigg| \right\} .	
\end{equation}
Notice that the typical estimators consist of two parts, i.e., a log-likelihood-based minimization led by Eq.~\eqref{eq:LL} and a parametric spectral density estimator $\mathbf{I}_{T}$ described by Eq.~\eqref{eq:I}. The former provides a theoretical guarantee for statistical optimization, and the latter empirically estimates the underlying spectral density of the observations. However, the periodogram estimation $\mathbf{I}_T(\lambda)$ is not a consistent estimator for $f_X(\lambda)$ despite asymptotically unbiased. To ensure the consistency and asymptotic normality of $\widehat{\boldsymbol{d}}_{\mathrm{GSE}}$, one has to force an admissible spectral density $f_X(\lambda)$ of the observations and assume Gaussianity for the source process $\boldsymbol{\epsilon}_t$, detailed in Appendix~\ref{app:assumptions}. Besides, the discrete Fourier transformation can only obtain which frequency components are contained in the whole process but does not know the time when each component appears. Thereby, the conventional statistical estimators have inherent defects in processing non-stationary signals. To alleviate this issue, some researchers adapt the tapered periodogram estimation for $f_X(\lambda)$ , that is, the tapered estimator $\widehat{\boldsymbol{d}}_{\mathrm{TSE}}$ by replacing $\mathbf{I}_T(\lambda_j)$ in Eq.~\eqref{eq:GSE} with $\mathbf{I}_T'(\lambda_j)$ in Eq.~\eqref{eq:J'}. Since the bias reduces along with an augmentation of the variance~\citep{velasco1999:cosine}, the tapered periodogram led by Eq.~\eqref{eq:J'} is still an inconsistent estimator of the spectral density $f_X$, leading to the pendent entanglement of the Gaussianity assumption.

\bibliography{JMrefs}
\bibliographystyle{plainnat}

\end{document}